\documentclass{article} 
\usepackage[margin=1in]{geometry}
\usepackage{times}
\usepackage[numbers,sort&compress]{natbib}
\usepackage{authblk}

\usepackage{amsmath,amsfonts,bm}

\def\eqref#1{equation~\ref{#1}}

\def\1{\bm{1}}

\DeclareMathAlphabet{\mathsfit}{\encodingdefault}{\sfdefault}{m}{sl}
\SetMathAlphabet{\mathsfit}{bold}{\encodingdefault}{\sfdefault}{bx}{n}

\usepackage{amssymb}
\usepackage{mathtools}
\usepackage{bbm}
\usepackage{algorithm}
\usepackage{algorithmic}
\usepackage{wrapfig}
\usepackage{hyperref}

\usepackage{url}
\usepackage{graphicx}
\usepackage{threeparttable}
\usepackage{multirow}
\usepackage{booktabs}
\usepackage[table]{xcolor}
\usepackage{fvextra}

\DefineVerbatimEnvironment{PromptBlock}{Verbatim}{
    fontsize=\scriptsize,
    breaklines=true,
    breakanywhere=true,
    frame=single,
    framesep=2mm,
    xleftmargin=0.5em,
    xrightmargin=0.5em
}

\usepackage{tabularx}
\usepackage{array}
\usepackage{caption}
\definecolor{lightblue}{RGB}{232,238,247}
\definecolor{myred}{RGB}{220,35,35}
\definecolor{myblue}{RGB}{35,85,170}

\title{ADAPT for Science: A Multi-Agent System for Self-Evolving Scientific Optimization}

\author[1,2]{Zhaoran Feng}
\author[2,3]{Yuchen Ni}
\author[2,3]{Jianan Liu}
\author[3]{Lemao Liu}
\author[3,5]{Dejing Dou}
\author[2,4]{Kairui Feng}

\affil[1]{National Laboratory of Solid State Microstructures, School of Physics and\\
Collaborative Innovation Center of Advanced Microstructures,\\
Nanjing University, Nanjing 210093, China}

\affil[2]{Shanghai Innovation Institute, Shanghai 200231, China}

\affil[3]{College of Computer Science and Artificial Intelligence,\\
Fudan University, Shanghai 200433, China}
 
\affil[4]{Shanghai Research Institute for Intelligent Autonomous Systems,\\
Tongji University, Shanghai 201210, China}

\affil[5]{BEDI Cloud, Beijing, China}

\date{}

\begin{document}

\maketitle
\vspace{-1cm}

\begin{abstract}
Scientific discovery often requires identifying a small number of high-performing designs in vast, constrained search spaces. This is difficult when evaluations are costly, feasible solutions are sparse, parameters are strongly coupled, and physical constraints are strict. Recent AI for AI methods can automatically improve algorithms and optimization procedures. However, general purpose approaches are not designed around scientific priors, physical constraints, and limited evaluation budgets. We introduce \textbf{ADAPT for Science}, a multi-agent framework that treats scientific optimization as two coupled decisions: \emph{where to search} and \emph{how to search}. ADAPT uses historical designs to identify promising and diverse initial conditions, progressively concentrates computation on favorable search directions, and evolves the optimization procedure from execution feedback. We evaluate ADAPT on the design of gravitational wave detectors using a differentiable physics simulator and eight unseen detector topologies. Under matched scientific evaluation budgets, ADAPT achieves a mean loss of $0.204 \pm 0.101$, reducing loss by $54.9\%$ relative to the best classical optimizer and by $62.2\%$ relative to the best general purpose agentic evolution baseline. ADAPT achieves the lowest loss on all unseen topologies among the evaluated baselines. ADAPT provides a step toward scientific optimization systems that continuously learn from prior experience and their own search outcomes.
\end{abstract}

\section{Introduction}
\label{sec:introduction}

Scientific progress often depends on finding a small number of useful
solutions in vast and constrained design spaces. From drug discovery to
gravitational-wave detector design, each candidate can be costly to evaluate,
many parameters interact strongly, and feasible solutions may occupy only a
small fraction of the search space. The central challenge is therefore not
only to evaluate a candidate, but to decide what is worth trying next.
This challenge becomes acute when optimization must proceed under a limited
experimental or computational budget. Search performance depends not only on
local parameter updates, but also on where the search begins, which prior
experience is reused, which optimization procedure is applied, and how
resources are allocated. These higher-level decisions are often fixed in advance or made manually, despite their substantial influence on the final
solution.
Gravitational-wave detector design provides a representative example.
Detector sensitivity depends on many interacting continuous parameters subject
to strict physical constraints. A poor initialization or an unsuitable
optimization strategy can consume substantial computation before reaching a
useful solution. At the same time, previous detector designs provide valuable
experience that can guide future search. The problem is therefore not
optimization from scratch, but how to reuse prior scientific experience,
choose where to begin, and adapt the search strategy as new evidence becomes
available.

AI has begun to automate parts of this process. Bayesian optimization and
autonomous laboratories use previous evaluations to select promising
candidates~\citep{shahriari2015taking,szymanski2023autonomous}, while
scientific agents automate planning and experimental
workflows~\citep{boiko2023autonomous}. These approaches primarily adapt the
scientific candidate while the higher-level search procedure remains largely
specified in advance. In parallel, recent program-search methods show that
language models can improve algorithms through evaluator
feedback~\citep{romera2024mathematical,shojaee2025llm}. These methods adapt the
procedure itself, but do not explicitly model reusable scientific experience,
target-dependent initialization, physical feasibility, and limited scientific
evaluation budgets.

We address this gap with \textbf{ADAPT for Science}, a multi-agent framework
for self-evolving scientific optimization. ADAPT treats the complete
scientific search policy as the object of optimization and decomposes it into
two coupled decisions: \emph{where to search} and \emph{how to search}.
Historical scientific designs provide offline experience for selecting
promising and diverse starting points and allocating evaluation effort.
Online execution feedback is then used to evolve the optimization procedure
itself. The resulting closed loop jointly adapts initialization, resource
allocation, and optimization strategy to the scientific target.
We evaluate ADAPT on gravitational-wave detector design using a differentiable
physics simulator and eight unseen detector topologies. Under matched
scientific evaluation budgets, ADAPT achieves a mean loss of
$0.204\pm0.101$, compared with $0.452\pm0.150$ for the strongest classical
optimizer and $0.540\pm0.263$ for the strongest general-purpose agentic
evolution baseline. These results correspond to reductions in mean loss of
\textbf{54.9\%} and \textbf{62.2\%}, respectively. ADAPT also achieves the lowest loss on all eight held-out topologies among the evaluated baselines. The results indicate that
scientific optimization benefits from adapting the search policy itself rather
than optimizing scientific parameters or optimization programs in isolation.

Our contributions are threefold:
\begin{itemize}

    \item We formulate scientific optimization at the level of the complete
    search policy rather than the scientific parameters alone, and decompose
    this policy into two coupled decisions: \emph{where to search} and
    \emph{how to search}.

    \item We develop \textbf{ADAPT for Science}, which combines reusable
    scientific experience with online execution feedback to jointly adapt
    initialization, resource allocation, and optimization procedures under
    physical and evaluation-budget constraints.

    \item We establish a held-out evaluation on eight unseen
    gravitational-wave detector topologies. ADAPT reduces mean loss by
    \textbf{54.9\%} relative to the strongest classical optimizer and by
    \textbf{62.2\%} relative to the strongest general-purpose agentic
    evolution baseline under matched scientific evaluation budgets, while
    achieving the lowest loss on all eight topologies among the evaluated baselines.

\end{itemize}

\section{Related Work}
\label{sec:related_work}

Scientific optimization is often formulated as sequential decision-making under
expensive evaluations and limited budgets.
Bayesian optimization and active learning leverage previous observations to
select informative evaluations
~\citep{shahriari2015taking,ren2023autonomous,zhang2023active,
jones1998efficient,snoek2012practical,gardner2014bayesian,
eriksson2019scalable}.
These methods have been widely applied to scientific discovery, including
chemical reaction optimization and autonomous experimentation
~\citep{hase2018phoenics,shields2021bayesian,
granda2018controlling,coley2019robotic,burger2020mobile,
macleod2020self}.
Recent scientific foundation models and autonomous agents further exploit
historical data, planning, and tool use for complex scientific workflows
~\citep{merchant2023scaling,wang2023scientific,boiko2023autonomous,
m2024augmenting,chen2026agentic}.
However, initialization, optimizer choice, and resource allocation are usually
specified beforehand.

A complementary direction treats the optimization algorithm itself as the search
object.
Learned optimizers and symbolic program search demonstrate that optimization
rules can be discovered from experience
~\citep{andrychowicz2016learning,chen2023symbolic,feng2026complexity,zhou2026prism}.
Algorithm discovery has further achieved improved computational procedures for
tasks such as matrix multiplication and sorting
~\citep{fawzi2022discovering,mankowitz2023faster}.
More recently, large language models enable program and heuristic evolution
through evaluator feedback, including FunSearch, EoH, ReEvo, and MCTS-AHD
~\citep{romera2024mathematical,liu2024evolution,ye2024reevo,
zheng2025monte}.
Related systems extend this idea to scientific equations, computational
workflows, and automated optimizer evolution
~\citep{shojaee2025llm,cheng2024trace,agrawal2026gepa,
lange2026shinkaevolve,shi2026generalizable,huang2026calm,
sharma2025openevolve}.
These approaches show that algorithms themselves can be optimized, but usually
without target-specific scientific priors.

ADAPT connects these directions by treating the complete scientific search
policy as the object of adaptation.
Historical designs provide priors for \emph{where to search}, while execution
feedback evolves \emph{how to search}.
By jointly adapting initialization, resource allocation, and optimization
dynamics, ADAPT enables efficient scientific optimization under physical
constraints and limited budgets.

\begin{figure*}
    \centering
    \includegraphics[width=\textwidth]{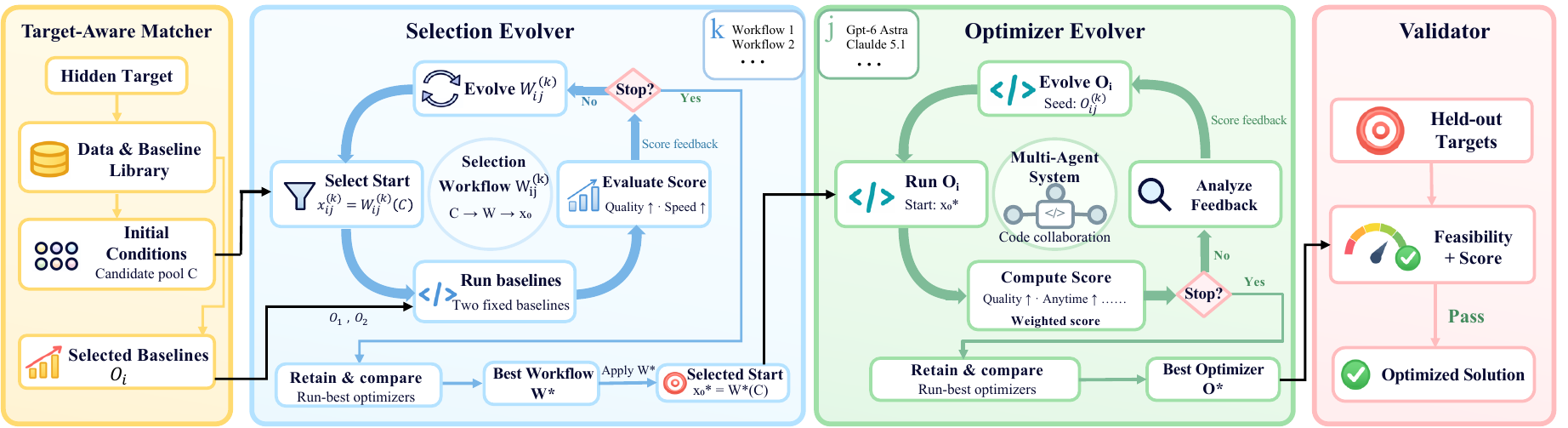}
    \caption{
    \textbf{Overview of ADAPT for Science.}
    A target-aware matcher retrieves historical designs and reference optimizers to form a target-specific search prior.
    The selection evolver determines \emph{where to search}, while the optimizer evolver determines \emph{how to search} from execution feedback.
    Together, they define a target-specific search policy that is evaluated on held-out targets under the original scientific objective and physical constraints.
    }
    \vspace{-10pt}
    \label{fig:adapt_overview}
\end{figure*}

\section{ADAPT FOR SCIENCE}
\label{sec:method}

Figure~\ref{fig:adapt_overview} summarizes ADAPT. Rather than optimizing only scientific variables, ADAPT optimizes the complete search policy through two coupled decisions, where to search and how to search. Historical experience provides a target-specific prior, online execution feedback adapts the optimization procedure, and an independent validator evaluates the resulting policy on unseen targets.

\subsection{Problem Formulation}
\label{sec:formulation}

For a scientific target $\tau\in\mathcal{T}$, let
$x\in\mathcal{X}_\tau$ denote the design variables,
$L_\tau(x)$ the objective, and
$\mathcal{F}_\tau\subseteq\mathcal{X}_\tau$ the feasible region.
Under a finite evaluation budget, performance depends on both the
initialization and the optimization procedure. We therefore define the
search policy as $\pi=(W,O)$, where $W$ selects the initial design
$x_0$ and $O$ generates the trajectory
$\mathcal{T}_\pi=\{x_0,\ldots,x_T\}$. For budget $B$, we define
$J_\tau(\pi;B)=\min_{x_t\in\mathcal{T}_\pi\cap\mathcal{F}_\tau,\,
c_t\leq B}L_\tau(x_t)$, where $c_t$ is the cumulative evaluation cost.

Conventional optimization fixes $W$ and $O$ in advance, whereas ADAPT
adapts both to each scientific target,
\begin{equation}
    (W_\tau^*,O_\tau^*)
    =
    \arg\min_{W\in\mathcal{W},\,O\in\mathcal{O}}
    J_\tau((W,O);B),
    \label{eq:adapt_objective}
\end{equation}
where $\mathcal{W}$ and $\mathcal{O}$ denote the spaces of selection
policies and optimization procedures. $W_\tau^*$ determines where to
search, while $O_\tau^*$ determines how to search.

\subsection{Historical Scientific Experience}
\label{sec:experience_library}

We construct a reusable experience library from historical
gravitational-wave detector designs and optimization records.
After cleaning, representation standardization, and transfer
compatibility filtering, the design library contains 15,409 valid
topology--design records, each pairing one topology with a compatible
parameterized design and its evaluated performance. A separate
baseline library contains ten established optimization procedures and
their historical behavior. The former supplies candidate
initializations, while the latter provides reference search behavior
for policy evolution. Held-out targets and their outcomes are excluded
from library construction and all subsequent strategy development.

\subsection{Target-Aware Retrieval and Initialization}
\label{sec:historical_retrieval}

For each target topology, ADAPT retrieves $K=64$ distinct historical
sources using a fixed structure-aware distance. A size-$3$ topology
contains nine interior symbols $\mathbf{c}$ and twelve boundary
symbols $\mathbf{b}$. For target $T$ and source $S$,
\begin{equation}
    d(T,S)
    =
    \frac{1}{9}\sum_{i=1}^{9}\delta(c_i^T,c_i^S)
    +
    \frac{1}{12}\sum_{j=1}^{12}
    \mathbf{1}\!\left[b_j^T\neq b_j^S\right],
    \label{eq:retrieval_distance}
\end{equation}
where $\delta(a,b)=0$ for identical symbols, $0.25$ for different
orientations within the same component family, and $1$ for different
families. Symbols A--D denote beamsplitters and E--H directional
beamsplitters, with four orientations per family. Symbols are compared
at fixed positions without rotational or reflection alignment.

Sources with the same complete topology as the target are excluded.
The remainder are ranked by increasing $d(T,S)$, with lexicographic
ordering used for ties. Incompatible records are skipped and the first
64 eligible sources are retained, without dataset-specific quotas.

Parameters are transferred by exact semantic matching of component
names, types, parameter properties, and complete coupling groups.
Matched bounded values are copied directly. Unmatched parameters use
frozen training-set medians in the order component-type and property,
property only, and global prior, followed by mapping to target bounds.
For parameter $x_i^{(k)}$ with bounds $\ell_i$ and $h_i$,
$u_i^{(k)}=\operatorname{clip}((x_i^{(k)}-\ell_i)/(h_i-\ell_i),
\epsilon,1-\epsilon)$ and
$z_i^{(k)}=\log(u_i^{(k)}/(1-u_i^{(k)}))$, with
$\epsilon=10^{-9}$. The resulting
$C_\tau=\{\mathbf{z}^{(k)}\}_{k=1}^{64}$ forms the candidate
initialization pool.

\subsection{Selection Evolution}
\label{sec:selection_evolution}

Given $C_\tau$, the Selection Evolver searches for a policy $W_\tau$
that returns one initialization under a fixed scientific evaluation
budget. Evolution starts from a direct 64-to-1 baseline that evaluates all
64 candidates and selects the best single initialization. Intermediate
candidate counts, screening stages, pruning rules, and budget-allocation
strategies are not predefined and are left to evolution.

A candidate policy may use historical attributes, transferred
parameters, target evaluations, feasibility, constraint violations,
and observed optimization progress to determine which candidates are
evaluated, retained, or discarded and how budget is allocated among
them. ADAPT evolves these policies with GEPA, OpenEvolve,
ShinkaEvolve, and ASI using GPT-6 Astra and Claude-fable-5.1.
Each workflow--model pair is compared with two reference baselines,
yielding 16 configurations.

At iteration $k$, a policy $W^{(k)}$ is executed under the fixed
budget, its trace is converted into feedback $F_W^{(k)}$, and the
feedback is used to propose $W^{(k+1)}$. Strong previous policies are
retained for direct comparison. Selection is based on measured
scientific performance, and the final policy produces
$x_{0,\tau}^*=W_\tau^*(C_\tau)$.

\subsection{Optimizer Evolution}
\label{sec:optimizer_evolution}

Starting from $x_{0,\tau}^*$, the Optimizer Evolver searches over
executable optimization programs evaluated directly by the
differentiable physics simulator under the same budget and numerical
constraints. The search space includes update rules, gradient and
state handling, learning-rate and momentum schedules, trajectory
management, numerical safeguards, and allocation of the remaining
evaluation budget.

The same workflow--model pool proposes and revises optimizers.
At iteration $k$, $O^{(k)}$ is executed, its trace is converted into
feedback $F_O^{(k)}$, and the feedback is used to generate
$O^{(k+1)}$. Strong previous optimizers remain available for direct
comparison. Candidate procedures are ranked by
\begin{equation}
    S
    =
    w_q S_{\mathrm{quality}}
    +
    w_a S_{\mathrm{anytime}}
    +
    w_f S_{\mathrm{feasibility}}
    +
    w_i S_{\mathrm{improvement}},
    \label{eq:optimizer_score}
\end{equation}
where the four terms measure final feasible quality, performance
throughout execution, physical validity, and improvement over the
initialization. This score is used only for strategy evolution.
Reported results use the native scientific objective and feasibility
criteria. The selected optimizer $O_\tau^*$ and selection policy
$W_\tau^*$ define $\pi_\tau^*=(W_\tau^*,O_\tau^*)$.

\subsection{Independent Validation}
\label{sec:validation}

The final policy $\pi_\tau^*$ is evaluated on detector topologies
absent from the experience library and excluded from retrieval,
baseline selection, policy evolution, and model selection. Their
outcomes are never returned to either evolver. Each held-out target is
evaluated under the same scientific budget as the comparison methods,
and all reported results use the native objective, physical
feasibility, and anytime optimization performance.

\section{Results}
\label{sec:results}

We first examine the policies discovered by the two evolution stages and then
evaluate their combined effect on eight held-out gravitational-wave detector
topologies. All comparisons use matched random seeds, hardware, numerical
precision, and the same total wall-clock budget. For ADAPT, this budget
covers both warmup (candidate screening and staged probing) and subsequent
optimization; no additional time is allocated for warmup.

\subsection{Selection evolution discovers progressive search}
\label{sec:results_selection}

\begin{figure*}[t]
\centering
\includegraphics[width=0.92\textwidth]{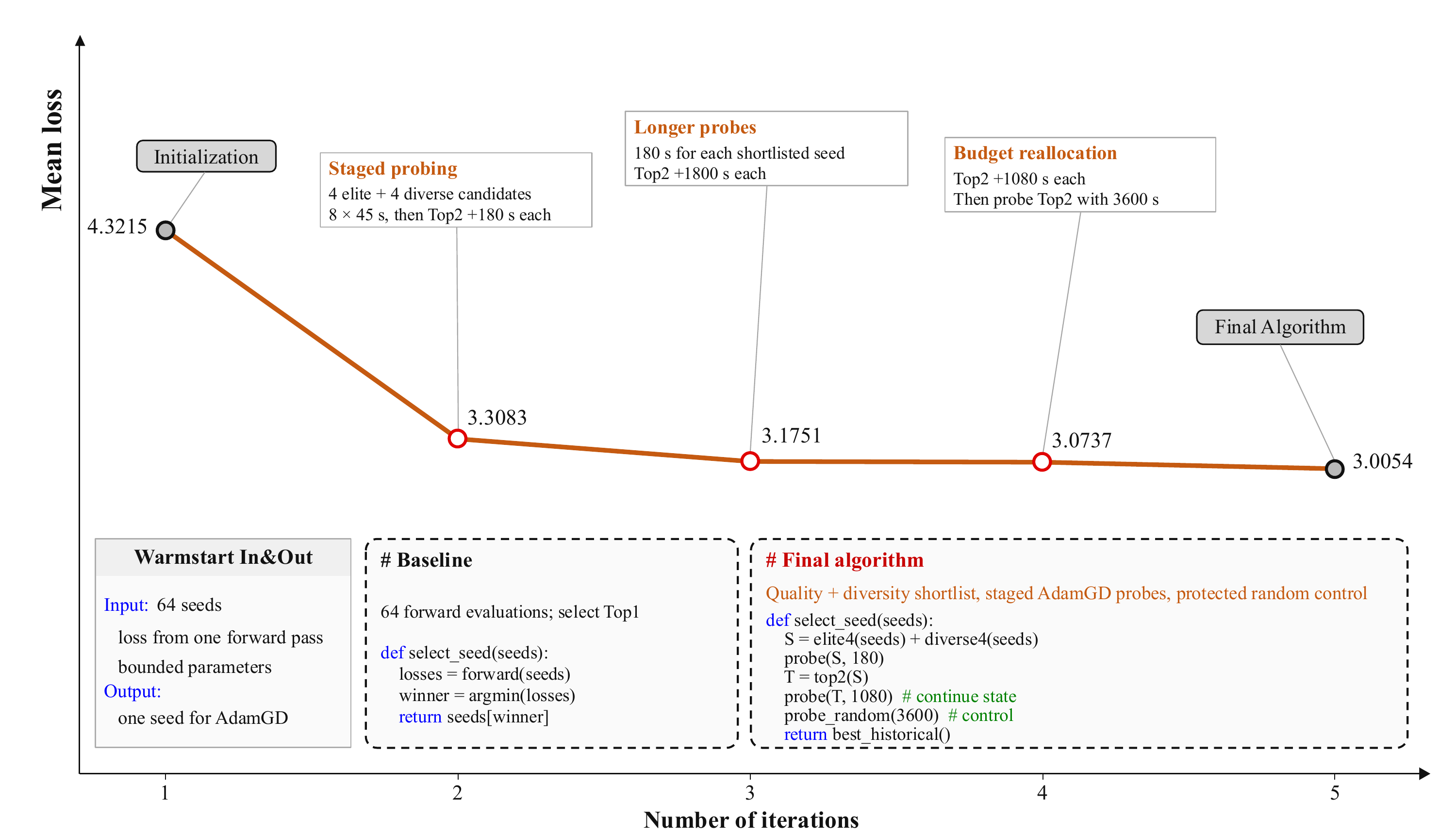}
\caption{
\textbf{Evolution of the selection policy.}
Mean loss across successive selection-policy iterations.
Starting from direct selection after 64 forward evaluations, evolution
introduces quality--diversity shortlisting, staged probing, and progressive
budget reallocation. The final policy reaches a mean loss of $3.0054$.
}
\vspace{-10pt}
\label{fig:selection_evolution_path}
\end{figure*}

\begin{wrapfigure}[16]{l}{0.6\columnwidth}
\centering
\vspace{-4pt}
\hspace{-1em}
\includegraphics[width=\linewidth]{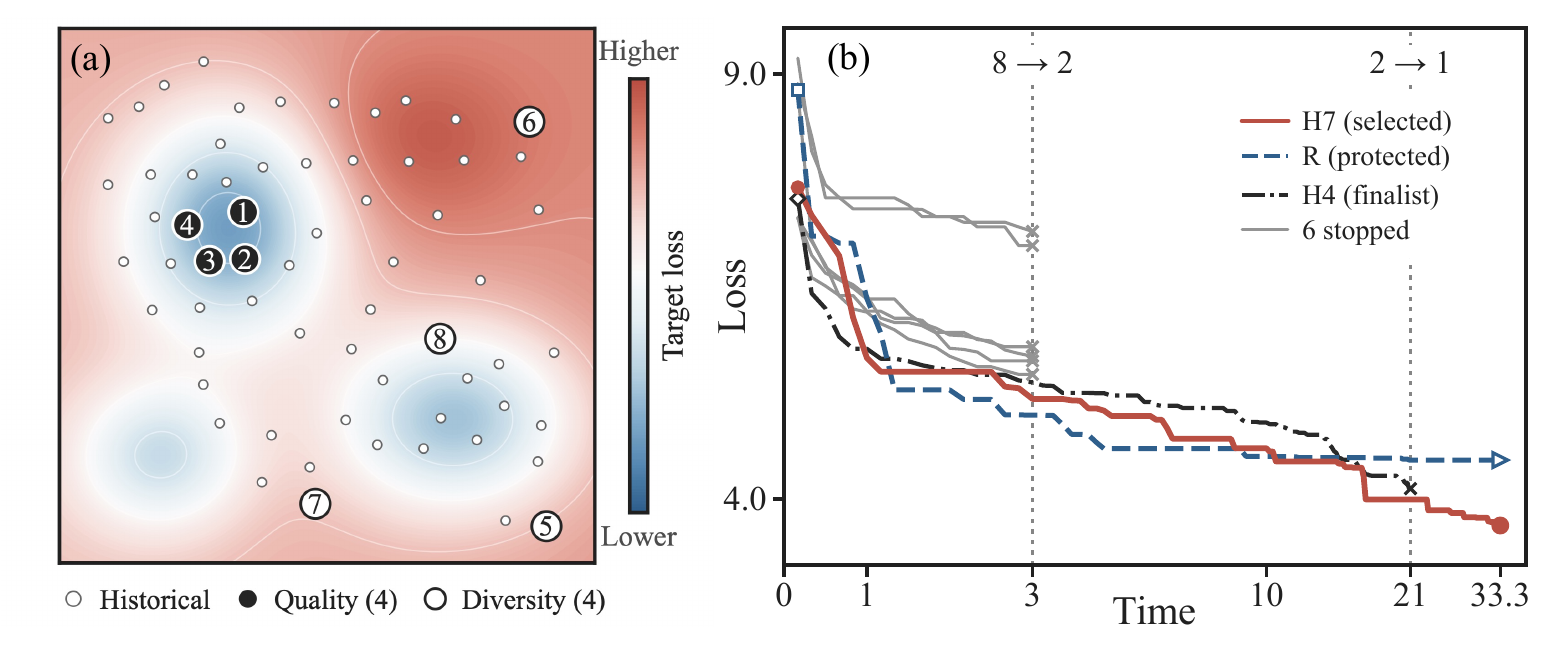}
\caption{
\textbf{Selection evolution discovers progressive search.}
Starting from a $64$-to-$1$ objective, the evolved policy selects four
high-quality feasible candidates and four diversity-preserving candidates,
then progressively concentrates evaluation as
$8\!\rightarrow\!2\!\rightarrow\!1$ while retaining protected exploration.
}
\label{fig:selection_results}
\vspace{-8pt}
\end{wrapfigure}
The Selection Evolver is given $64$ transferred initializations and a single
objective, to return one start under a fixed evaluation budget. No intermediate
candidate count, pruning schedule, or budget-allocation rule is specified.
Figure~\ref{fig:selection_evolution_path} shows how the policy develops from
direct Top-1 selection into a staged search procedure. Successive iterations
introduce quality--diversity shortlisting, longer target-specific probes, and
progressive budget reallocation, reducing the mean loss to
$3.0054$. The resulting policy converges to the
$64\!\rightarrow\!8\!\rightarrow\!2\!\rightarrow\!1$
workflow shown in Figure~\ref{fig:selection_results}.

The first stage evaluates all $64$ starts once on the target without optimizer
updates and records loss, feasibility, and constraint violation. Among feasible
candidates, the four lowest-loss starts form an elite set. Four additional
starts are then selected for coverage in transferred parameter space. For
candidate $k$, each bounded parameter is normalized as
$q_{k,i}=(x_{k,i}-\ell_i)/(h_i-\ell_i)$, and pairwise distance is
$D(k,j)=\sqrt{d^{-1}\sum_i(q_{k,i}-q_{j,i})^2}$. Starting from the four
elites, each additional candidate is chosen by
$k^\star=\arg\max_{k\in\mathcal{F}\setminus S}\min_{j\in S}D(k,j)$
until eight starts are retained. Ties are resolved by candidate index.
If fewer than eight feasible starts exist, all feasible starts are retained
and numerically valid infeasible candidates are added in order of constraint
violation, loss, and candidate index. Fewer than eight numerically valid
candidates terminates the selection run.

The eight retained starts then enter a target-specific race. Each receives a
$3$-minute optimization run, after which two leading trajectories are retained
and continued to $21$ minutes. A protected exploratory branch is maintained
during pruning, and the remaining budget is concentrated on the final start.
The discovered policy therefore moves from historical quality and parameter
coverage to target-specific execution evidence before committing the remaining
budget to a single direction.



\subsection{Optimizer evolution discovers budget-aware search dynamics}
\label{sec:results_optimizer}
\begin{figure*}
    \centering
    \includegraphics[width=0.9\textwidth]{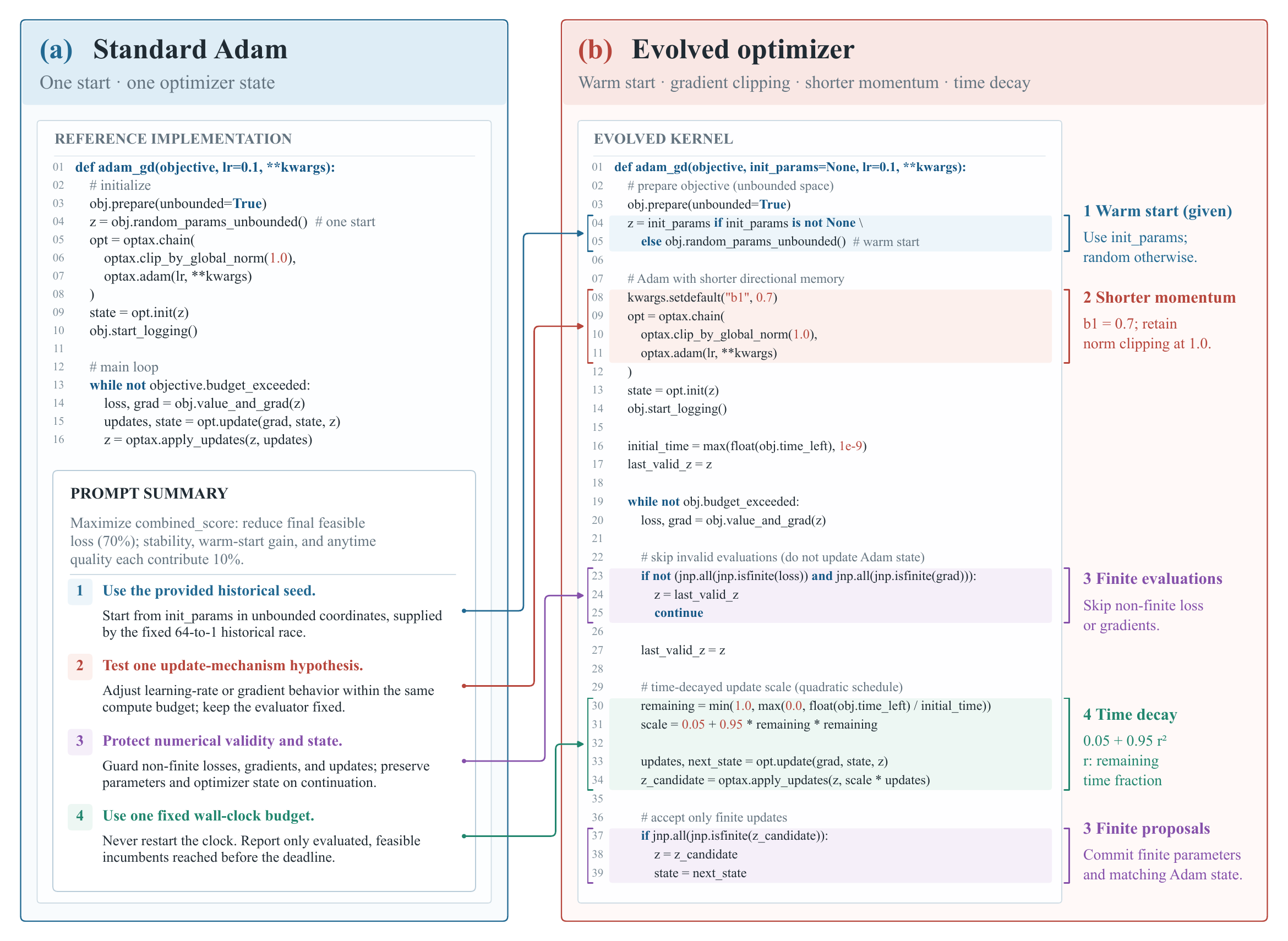}
    \caption{
    \textbf{Optimizer evolution discovers budget-aware search dynamics.}
    The evolved procedure retains the Adam structure while using the selected
    initialization, shorter first-moment memory, time-dependent updates, and
    explicit rejection of invalid evaluations and proposals under a fixed
    scientific budget.
    }
    \vspace{-14pt}
    \label{fig:optimizer_results}
\end{figure*}

\begin{figure*}[t]
\centering
\includegraphics[width=0.9\textwidth]{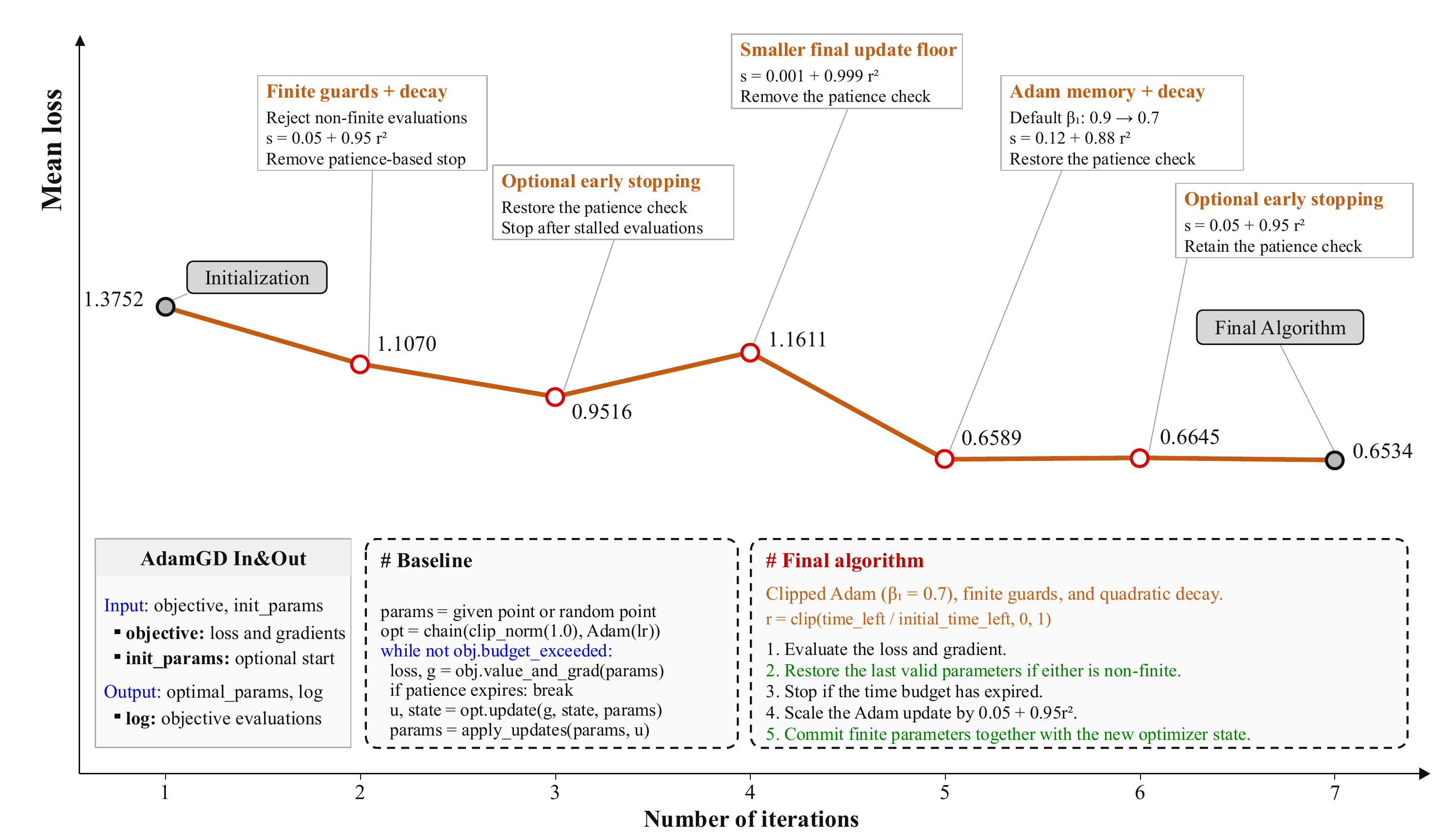}
\caption{
\textbf{Evolution of the optimization procedure.}
Mean loss across successive optimizer-evolution iterations, with annotations
summarizing the changes tested at each stage. Starting from the reference
AdamGD implementation, evolution tests numerical safeguards, stopping
behavior, update scaling, and optimizer memory. The final procedure reaches
a mean loss of $0.6534$ during strategy evolution.
}
\vspace{-14pt}
\label{fig:optimizer_evolution_path}
\end{figure*}

The preceding selection stage supplies the initialization, while the Optimizer
Evolver determines how the subsequent local search proceeds.
Figure~\ref{fig:optimizer_evolution_path} traces this process from the
reference AdamGD implementation to the final optimization procedure.
Successive iterations test finite-state protection, time-dependent update
scaling, stopping behavior, and shorter optimizer memory. These changes are
retained or revised according to measured execution performance, ultimately
yielding the optimizer shown in Figure~\ref{fig:optimizer_results}.

Figure~\ref{fig:optimizer_results} shows the resulting optimizer selected by
ADAPT. Rather than replacing Adam, evolution retains its basic structure and
modifies its behavior under the finite evaluation horizon.

The discovered procedure shortens the first-moment memory to
$\beta_1=0.7$ while retaining global gradient clipping. It further scales each
update by $s(r)=0.05+0.95r^2$, where $r$ is the remaining fraction of the
evaluation budget, producing larger updates early and increasingly conservative
updates near the deadline. Non-finite losses or gradients are skipped, and
non-finite parameter proposals are rejected together with the corresponding
optimizer-state update.

These changes couple the local search to both the selected historical start and
the remaining evaluation horizon. The resulting optimizer remains close to
Adam computationally, but adapts its directional memory, update magnitude, and
numerical state handling to the observed scientific search process.


\subsection{Overall Performance on Held-Out Topologies}
\label{sec:overall_results}

We compare the complete ADAPT policy with fixed classical optimizers and
general-purpose agentic evolution methods on eight held-out detector
topologies. Lower loss indicates better performance.
\begin{table*}[t]
\centering

\caption{
\textbf{Optimization performance of classical optimizers across eight held-out
gravitational-wave detector targets.}
Lower loss is better.
Red percentages indicate the relative increase in loss compared with Ours.
The final column reports mean $\pm$ standard deviation.
}
\label{tab:classical_results}

\definecolor{oursblue}{RGB}{240,241,255}

\newcommand{\rel}[1]{%
    \hspace{0.4pt}{\color{red}\tiny #1\%}%
}

\resizebox{\textwidth}{!}{
\setlength{\tabcolsep}{1.5pt}
\renewcommand{\arraystretch}{1.10}

\begin{tabular}{@{}lccccccccc@{}}
\toprule

\textbf{Method}
& \textbf{T1}
& \textbf{T2}
& \textbf{T3}
& \textbf{T4}
& \textbf{T5}
& \textbf{T6}
& \textbf{T7}
& \textbf{T8}
& \textbf{Mean $\pm$ Std.}
\\

\midrule

\textbf{Adam}
& 0.423 \rel{+36.0}
& 0.887 \rel{+117.4}
& 1.833 \rel{+2344.0}
& 0.332 \rel{+45.6}
& 0.379 \rel{+152.7}
& 0.692 \rel{+397.8}
& 0.993 \rel{+509.2}
& 0.666 \rel{+329.7}
& $0.776\pm0.458$ \rel{+280.4}
\\

\textbf{NAAdam}
& 0.423 \rel{+36.0}
& 0.696 \rel{+70.6}
& 0.423 \rel{+464.0}
& 0.334 \rel{+46.5}
& 0.415 \rel{+176.7}
& 0.375 \rel{+169.8}
& 0.695 \rel{+326.4}
& 0.255 \rel{+64.5}
& $0.452\pm0.150$ \rel{+121.6}
\\

\textbf{Lion}
& 4.358 \rel{+1301.3}
& 3.841 \rel{+841.4}
& 2.028 \rel{+2604.0}
& 3.028 \rel{+1228.1}
& 2.004 \rel{+1236.0}
& 2.641 \rel{+1800.0}
& 2.663 \rel{+1533.7}
& 2.325 \rel{+1400.0}
& $2.861\pm0.793$ \rel{+1302.5}
\\

\textbf{SGDM}
& 4.151 \rel{+1234.7}
& 0.445 \rel{+9.1}
& 0.468 \rel{+524.0}
& 0.473 \rel{+107.5}
& 0.581 \rel{+287.3}
& 0.448 \rel{+222.3}
& 0.196 \rel{+20.2}
& 0.415 \rel{+167.7}
& $0.897\pm1.234$ \rel{+339.7}
\\

\textbf{Sophia}
& 0.506 \rel{+62.7}
& 0.412 \rel{+1.0}
& 0.627 \rel{+736.0}
& 1.749 \rel{+667.1}
& 0.653 \rel{+335.3}
& 0.574 \rel{+312.9}
& 0.430 \rel{+163.8}
& 1.659 \rel{+970.3}
& $0.826\pm0.514$ \rel{+304.9}
\\

\textbf{Adan}
& 1.404 \rel{+351.4}
& 1.321 \rel{+223.8}
& 1.376 \rel{+1734.7}
& 2.783 \rel{+1120.6}
& 2.212 \rel{+1374.7}
& 0.673 \rel{+384.2}
& 1.355 \rel{+731.3}
& 1.387 \rel{+794.8}
& $1.564\pm0.602$ \rel{+666.7}
\\

\midrule

\rowcolor{oursblue}
\textbf{Ours}
& \textbf{0.311}
& \textbf{0.408}
& \textbf{0.075}
& \textbf{0.228}
& \textbf{0.150}
& \textbf{0.139}
& \textbf{0.163}
& \textbf{0.155}
& $\mathbf{0.204\pm0.101}$
\\

\bottomrule
\end{tabular}
}

\vspace{10pt}

\caption{
\textbf{Comparison with agentic evolution baselines across eight held-out
gravitational-wave detector targets.}
Lower loss is better.
Red percentages indicate the relative increase in loss compared with Ours.
The final column reports mean $\pm$ standard deviation.
}
\label{tab:agent_results}

\newcommand{\relagent}[1]{%
    \hspace{0.4pt}{\color{red}\tiny #1\%}%
}

\resizebox{\textwidth}{!}{
\setlength{\tabcolsep}{2.0pt}
\renewcommand{\arraystretch}{1.10}

\begin{tabular}{@{}lccccccccc@{}}
\toprule

\textbf{Method}
& \textbf{T1}
& \textbf{T2}
& \textbf{T3}
& \textbf{T4}
& \textbf{T5}
& \textbf{T6}
& \textbf{T7}
& \textbf{T8}
& \textbf{Mean $\pm$ Std.}
\\

\midrule

\textbf{GEPA}
& 0.557 \relagent{+79.1}
& 0.410 \relagent{+0.5}
& 0.480 \relagent{+540.0}
& 1.218 \relagent{+434.2}
& 0.379 \relagent{+152.7}
& 0.435 \relagent{+212.9}
& 0.475 \relagent{+191.4}
& 0.365 \relagent{+135.5}
& $0.540\pm0.263$ \relagent{+164.7}
\\

\textbf{OpenE.}
& 0.707 \relagent{+127.3}
& 0.492 \relagent{+20.6}
& 0.449 \relagent{+498.7}
& 0.261 \relagent{+14.5}
& 0.665 \relagent{+343.3}
& 0.426 \relagent{+206.5}
& 0.439 \relagent{+169.3}
& 0.954 \relagent{+515.5}
& $0.549\pm0.202$ \relagent{+169.1}
\\

\textbf{ShinkaE.}
& 0.561 \relagent{+80.4}
& 0.440 \relagent{+7.8}
& 0.674 \relagent{+798.7}
& 2.632 \relagent{+1054.4}
& 0.366 \relagent{+144.0}
& 0.210 \relagent{+51.1}
& 0.643 \relagent{+294.5}
& 0.402 \relagent{+159.4}
& $0.741\pm0.729$ \relagent{+263.2}
\\

\textbf{ASI}
& 0.725 \relagent{+133.1}
& 0.483 \relagent{+18.4}
& 0.624 \relagent{+732.0}
& 0.653 \relagent{+186.4}
& 1.309 \relagent{+772.7}
& 0.450 \relagent{+223.7}
& 1.681 \relagent{+931.3}
& 1.067 \relagent{+588.4}
& $0.874\pm0.410$ \relagent{+328.4}
\\

\midrule

\rowcolor{oursblue}
\textbf{Ours}
& \textbf{0.311}
& \textbf{0.408}
& \textbf{0.075}
& \textbf{0.228}
& \textbf{0.150}
& \textbf{0.139}
& \textbf{0.163}
& \textbf{0.155}
& $\mathbf{0.204\pm0.101}$
\\

\bottomrule
\end{tabular}
}

\end{table*}

Among fixed optimizers, NAAdam gives the lowest mean loss at
$0.452\pm0.150$, whereas ADAPT reaches
$\mathbf{0.204\pm0.101}$, a \textbf{54.9\% reduction}.
ADAPT also achieves the lowest loss on all eight held-out topologies among the evaluated baselines, showing
that the mean improvement is not driven by a small subset of targets.

The same advantage persists against agentic evolution baselines. GEPA gives
the strongest mean result among these methods at $0.540\pm0.263$, compared
with $0.204\pm0.101$ for ADAPT, corresponding to a
\textbf{62.2\% reduction}. Thus, program evolution alone does not match the
full system under the present protocol. Combining target-specific historical
initialization with evolved selection and optimization policies yields the
stronger held-out performance.

\subsection{Representative Detector Optimization}
\label{sec:case_study}

\begin{figure*}[t]
    \centering
    \includegraphics[width=\textwidth]{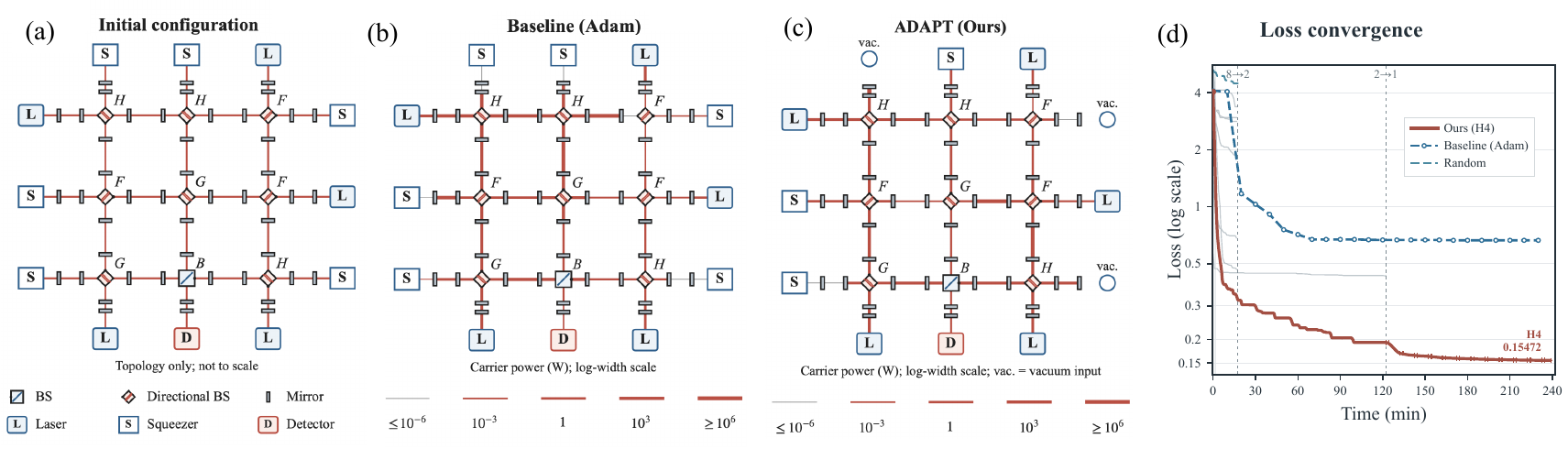}
    \caption{
    \textbf{Representative held-out detector optimization.}
    (a) Initial detector configuration.
    (b) Design obtained with standard Adam and its optimized carrier-power
    distribution.
    (c) Final ADAPT design.
    ADAPT suppresses three optical branches with near-zero optimized
    contributions and replaces them with vacuum inputs while preserving the
    remaining detector structure.
    (d) Loss convergence under the same scientific evaluation budget.
    ADAPT reaches a final reported loss of $0.155$.
    }
    \label{fig:target_case}
\end{figure*}

Figure~\ref{fig:target_case} shows how the numerical gain appears in a
representative held-out detector. Standard Adam largely preserves the initial
structure while adjusting continuous parameters. ADAPT instead suppresses
three optical branches whose optimized contributions approach zero, yielding
a simpler effective configuration while preserving the remaining optical
structure. Under the same evaluation budget, Adam reaches a loss of $0.666$
and ADAPT reaches $0.155$, a \textbf{76.7\% reduction}. The performance gain
therefore coincides with a distinct, physically interpretable change in the
optimized detector rather than a small perturbation of the baseline solution.
Additional held-out examples are provided in Appendix E.

\subsection{Ablation of Where and How to Search}
\label{sec:ablation}
\begin{wraptable}[12]{l}{0.55\columnwidth}
    \centering
    \vspace{-14pt}
    \captionsetup{skip=8pt}

    \caption{
    \textbf{Ablation study.}
Mean $\pm$ std.\ over eight held-out topologies (lower is better).
    }
    \label{tab:ablation}

    \definecolor{oursblue}{RGB}{240,241,255}

    \newcommand{\relabl}[1]{%
        \hspace{0.4pt}{\color{red}\fontsize{7}{8}\selectfont #1\%}%
    }

    \scriptsize
    \setlength{\tabcolsep}{0.8pt}
    \renewcommand{\arraystretch}{1.08}

    \begin{tabularx}{\linewidth}{
        @{}
        l
        >{\centering\arraybackslash}X
        >{\centering\arraybackslash}X
        >{\centering\arraybackslash}X
        >{\centering\arraybackslash}X
        @{}
    }
        \toprule

        \textbf{Target}
        & \textbf{Direct+GEPA}
        & \textbf{Fixed+Adam}
        & \textbf{No-Prompt}
        & \cellcolor{oursblue}\textbf{Ours}
        \\

        \midrule

        T1
        & 0.416 \relabl{+33.8}
        & 0.320 \relabl{+2.9}
        & 0.424 \relabl{+36.3}
        & \cellcolor{oursblue}\textbf{0.311}
        \\

        T2
        & 0.986 \relabl{+141.7}
        & 0.371 \relabl{-9.1}
        & 0.443 \relabl{+8.6}
        & \cellcolor{oursblue}\textbf{0.408}
        \\

        T3
        & 0.373 \relabl{+397.3}
        & 0.180 \relabl{+140.0}
        & 0.145 \relabl{+93.3}
        & \cellcolor{oursblue}\textbf{0.075}
        \\

        T4
        & 0.476 \relabl{+108.8}
        & 0.246 \relabl{+7.9}
        & 0.287 \relabl{+25.9}
        & \cellcolor{oursblue}\textbf{0.228}
        \\

        T5
        & 0.164 \relabl{+9.3}
        & 0.200 \relabl{+33.3}
        & 0.264 \relabl{+76.0}
        & \cellcolor{oursblue}\textbf{0.150}
        \\

        T6
        & 0.138 \relabl{-0.7}
        & 0.143 \relabl{+2.9}
        & 0.148 \relabl{+6.5}
        & \cellcolor{oursblue}\textbf{0.139}
        \\

        T7
        & 0.413 \relabl{+153.4}
        & 0.415 \relabl{+154.6}
        & 0.499 \relabl{+206.1}
        & \cellcolor{oursblue}\textbf{0.163}
        \\

        T8
        & 0.421 \relabl{+171.6}
        & 0.172 \relabl{+11.0}
        & 0.210 \relabl{+35.5}
        & \cellcolor{oursblue}\textbf{0.155}
        \\

        \midrule

        \textbf{Mean}
        & \shortstack{
            $0.423\pm$\\
            $0.243$ \relabl{+107.4}
          }
        & \shortstack{
            $0.256\pm$\\
            $0.094$ \relabl{+25.5}
          }
        & \shortstack{
            $0.303\pm$\\
            $0.128$ \relabl{+48.5}
          }
        & \cellcolor{oursblue}
          \shortstack{
            $\mathbf{0.204\pm}$\\
            $\mathbf{0.101}$
          }
        \\

        \bottomrule
    \end{tabularx}

    \vspace{-6pt}
\end{wraptable}

We isolate the contributions of adaptive initialization, optimizer evolution,
and science-specific guidance. \textbf{Direct+GEPA} bypasses evolved start
selection and applies GEPA from a single retrieved initialization.
\textbf{Fixed+Adam} retains the selected initialization but replaces the
evolved optimizer with fixed Adam. \textbf{No-Prompt} removes the
science-specific guidance while retaining the corresponding agentic evolution
pipeline.

Removing evolved start selection increases the mean loss to
$0.423\pm0.243$, and full ADAPT reduces this value by
\textbf{51.8\%}. ADAPT is lower on seven of eight targets, with T6
effectively unchanged at $0.138$ versus $0.139$. Retaining the selected start
but replacing the evolved optimizer with Adam gives $0.256\pm0.094$.
ADAPT reduces this loss by \textbf{20.3\%} and is lower on seven targets, with
T2 as the exception. Removing science-specific guidance gives
$0.303\pm0.128$, whereas full ADAPT is \textbf{32.7\%} lower and improves all
eight targets.

These ablations separate the two adaptive decisions in ADAPT. Start selection
reduces sensitivity to initialization, optimizer evolution improves subsequent
local search, and science-specific guidance strengthens the evolution process
across held-out targets.

\section{Limitations and Future Work}
\label{sec:limitations}

ADAPT is designed for scientific optimization with reusable historical
experience, expensive evaluations, and explicit physical constraints.
The present study focuses on gravitational-wave detector design, where a
differentiable simulator provides controlled evaluation and a large historical
design library.
The results therefore establish the effectiveness of the framework in one
challenging scientific domain, but do not yet demonstrate universality across
scientific disciplines.

The method also depends on the quality and coverage of its experience library.
Nearest-neighbor retrieval is effective when historical designs contain useful
local structure, but its advantage may diminish for targets that lie far
outside the observed design distribution.
In addition, evolving both selection workflows and optimization programs
introduces computational overhead relative to a single fixed optimizer.
This overhead is justified when scientific evaluations are sufficiently costly,
but should be reduced for cheaper optimization problems.

Several extensions follow naturally.
First, ADAPT can be evaluated on scientific problems with different search
spaces, simulators, and physical constraints, including non-differentiable and
experimental settings.
Second, the experience library can be updated online so that successful
solutions and optimizers become reusable knowledge for future targets.
Third, more efficient strategy evolution could reduce the number of
workflow--model evaluations through learned transfer, uncertainty-aware
allocation, or meta-optimization.
Finally, extending the independent validator to real experimental measurements
would test whether self-evolving optimization remains reliable beyond
simulation.

\section{Conclusion}
\label{sec:conclusion}

We introduced \textbf{ADAPT for Science}, a framework that treats the complete
scientific search policy as the object of optimization. ADAPT decomposes this
policy into two coupled decisions: \emph{where to search} and \emph{how to
search}. Historical scientific experience guides target-specific
initialization and resource allocation, while online execution feedback adapts
the optimization procedure itself.

On eight held-out gravitational-wave detector topologies, ADAPT achieves a
mean loss of $0.204\pm0.101$, reducing mean loss by \textbf{54.9\%} relative
to the strongest classical optimizer and by \textbf{62.2\%} relative to the
strongest general-purpose agentic evolution baseline under matched scientific
evaluation budgets. ADAPT also achieves the lowest loss on all eight held-out topologies among the evaluated baselines. Ablation results further show that adaptive initialization,
optimizer evolution, and science-specific guidance each contribute to the
full-system performance.

These results support a broader view of scientific automation in which the
search procedure itself is adaptive. Rather than optimizing scientific
parameters under a fixed workflow, ADAPT combines prior scientific experience
with execution feedback to continually refine how the search is initialized
and conducted.

\subsection*{AI use statement}

Large language models were used as components of the agentic optimization
workflows to propose and revise candidate search policies and optimization
procedures based on execution feedback. All candidates were evaluated by the
scientific simulator and objective functions.

Generative AI tools were used only for language editing and manuscript
polishing. The authors reviewed all AI-assisted edits and take full
responsibility for the final manuscript.

\subsection*{Reproducibility statement}

The supplementary material provides the scientific objective, data
construction, optimization procedures, evaluation protocol, implementation
settings, and additional experimental details needed to reproduce the study.
\bibliography{iclr2027_conference}

@article{wang2023scientific,
  title={Scientific discovery in the age of artificial intelligence},
  author={Wang, Hanchen and Fu, Tianfan and Du, Yuanqi and Gao, Wenhao and Huang, Kexin and Liu, Ziming and Chandak, Payal and Liu, Shengchao and Van Katwyk, Peter and Deac, Andreea and others},
  journal={Nature},
  volume={620},
  number={7972},
  pages={47--60},
  year={2023},
  publisher={Nature Publishing Group UK London}
}

@article{szymanski2023autonomous,
  title={An autonomous laboratory for the accelerated synthesis of inorganic materials},
  author={Szymanski, Nathan J and Rendy, Bernardus and Fei, Yuxing and Kumar, Rishi E and He, Tanjin and Milsted, David and McDermott, Matthew J and Gallant, Max and Cubuk, Ekin Dogus and Merchant, Amil and others},
  journal={Nature},
  volume={624},
  number={7990},
  pages={86--91},
  year={2023},
  publisher={Nature Publishing Group UK London}
}

@article{boiko2023autonomous,
  title={Autonomous chemical research with large language models},
  author={Boiko, Daniil A and MacKnight, Robert and Kline, Ben and Gomes, Gabe},
  journal={Nature},
  volume={624},
  number={7992},
  pages={570--578},
  year={2023},
  publisher={Nature Publishing Group UK London}
}

@article{romera2024mathematical,
  title={Mathematical discoveries from program search with large language models},
  author={Romera-Paredes, Bernardino and Barekatain, Mohammadamin and Novikov, Alexander and Balog, Matej and Kumar, M Pawan and Dupont, Emilien and Ruiz, Francisco JR and Ellenberg, Jordan S and Wang, Pengming and Fawzi, Omar and others},
  journal={Nature},
  volume={625},
  number={7995},
  pages={468--475},
  year={2024},
  publisher={Nature Publishing Group UK London}
}

@inproceedings{shojaee2025llm,
  title={Llm-sr: Scientific equation discovery via programming with large language models},
  author={Shojaee, Parshin and Meidani, Kazem and Gupta, Shashank and Barati Farimani, Amir and Reddy, Chandan},
  booktitle={International Conference on Learning Representations},
  volume={2025},
  pages={16054--16085},
  year={2025}
}

@article{shahriari2015taking,
  title={Taking the human out of the loop: A review of Bayesian optimization},
  author={Shahriari, Bobak and Swersky, Kevin and Wang, Ziyu and Adams, Ryan P and De Freitas, Nando},
  journal={Proceedings of the IEEE},
  volume={104},
  number={1},
  pages={148--175},
  year={2015},
  publisher={IEEE}
}

@article{ye2024reevo,
  title={Reevo: Large language models as hyper-heuristics with reflective evolution},
  author={Ye, Haoran and Wang, Jiarui and Cao, Zhiguang and Berto, Federico and Hua, Chuanbo and Kim, Haeyeon and Park, Jinkyoo and Song, Guojie},
  journal={Advances in neural information processing systems},
  volume={37},
  pages={43571--43608},
  year={2024}
}

@article{zheng2025monte,
  title={Monte carlo tree search for comprehensive exploration in llm-based automatic heuristic design},
  author={Zheng, Zhi and Xie, Zhuoliang and Wang, Zhenkun and Hooi, Bryan},
  journal={arXiv preprint arXiv:2501.08603},
  year={2025}
}

@article{liu2024evolution,
  title={Evolution of heuristics: Towards efficient automatic algorithm design using large language model},
  author={Liu, Fei and Tong, Xialiang and Yuan, Mingxuan and Lin, Xi and Luo, Fu and Wang, Zhenkun and Lu, Zhichao and Zhang, Qingfu},
  journal={arXiv preprint arXiv:2401.02051},
  year={2024}
}

@inproceedings{agrawal2026gepa,
  title={Gepa: Reflective prompt evolution can outperform reinforcement learning},
  author={Agrawal, Lakshya A and Tan, Shangyin and Soylu, Dilara and Ziems, Noah and Khare, Rishi and Opsahl-Ong, Krista and Singhvi, Arnav and Shandilya, Herumb and Ryan, Michael J and Jiang, Meng and others},
  booktitle={International Conference on Learning Representations},
  volume={2026},
  pages={8479--8565},
  year={2026}
}

@inproceedings{lange2026shinkaevolve,
  title={Shinkaevolve: Towards open-ended and sample-efficient program evolution},
  author={Lange, Robert and Imajuku, Yuki and Cetin, Edoardo},
  booktitle={International Conference on Learning Representations},
  volume={2026},
  pages={74026--74078},
  year={2026}
}

@article{sharma2025openevolve,
  title={OpenEvolve: an open-source evolutionary coding agent},
  author={Sharma, Asankhaya},
  journal={GitHub repository. Accessed},
  pages={08--10},
  year={2025}
}

@article{ren2023autonomous,
  title={Autonomous experiments using active learning and AI},
  author={Ren, Zhichu and Ren, Zekun and Zhang, Zhen and Buonassisi, Tonio and Li, Ju},
  journal={Nature Reviews Materials},
  volume={8},
  number={9},
  pages={563--564},
  year={2023},
  publisher={Nature Publishing Group UK London}
}

@inproceedings{zhou2026prism,
  title={PRISM: Probabilistic Reward Model with Inherent Structural Modeling},
  author={Zhou, Yuhang and Cao, Yixin and Ni, Yuchen and Dou, Shihan and Chen, Xutian and Zhang, Ge and Liu, Xiang and Ye, Guangnan},
  booktitle={Proceedings of the 64th Annual Meeting of the Association for Computational Linguistics (Volume 1: Long Papers)},
  pages={12345--12362},
  year={2026}
}

@article{feng2026complexity,
  title={Complexity-powered machine intelligent classification of quantum many-body dynamics},
  author={Feng, Zhaoran and Chen, Jiangzhi and Wang, Ce and Ren, Jie},
  journal={Physical Review E},
  volume={113},
  number={6},
  pages={065301},
  year={2026},
  publisher={APS}
}

@article{m2024augmenting,
  title={Augmenting large language models with chemistry tools},
  author={M. Bran, Andres and Cox, Sam and Schilter, Oliver and Baldassari, Carlo and White, Andrew D and Schwaller, Philippe},
  journal={Nature machine intelligence},
  volume={6},
  number={5},
  pages={525--535},
  year={2024},
  publisher={Nature Publishing Group UK London}
}

@article{chen2026agentic,
  title={An agentic artificially intelligent X-ray scientist},
  author={Chen, Zhantao and Petsch, Alexander N and Israelski, Aidan J and Plumley, Rajan and Shen, Lingjia and Wang, Cong and Peng, Cheng and Ni, Yuan and Bansil, Arun and Chowdhury, Sugata and others},
  journal={Nature Machine Intelligence},
  pages={1--12},
  year={2026},
  publisher={Nature Publishing Group UK London}
}

@article{zhang2023active,
  title={Active learning for optimal intervention design in causal models},
  author={Zhang, Jiaqi and Cammarata, Louis and Squires, Chandler and Sapsis, Themistoklis P and Uhler, Caroline},
  journal={Nature Machine Intelligence},
  volume={5},
  number={10},
  pages={1066--1075},
  year={2023},
  publisher={Nature Publishing Group UK London}
}

@inproceedings{shi2026generalizable,
  title={Generalizable heuristic generation through LLMs with meta-optimization},
  author={Shi, Yiding and Zhou, Jianan and Song, Wen and Bi, Jieyi and Wu, Yaoxin and Cao, Zhiguang and Zhang, Jie},
  booktitle={International Conference on Learning Representations},
  volume={2026},
  pages={119314--119349},
  year={2026}
}

@inproceedings{huang2026calm,
  title={Calm: Co-evolution of algorithms and language model for automatic heuristic design},
  author={Huang, Ziyao and Wu, Weiwei and Wu, Kui and Lee, Wei-Bin and Wang, Jianping},
  booktitle={International Conference on Learning Representations},
  volume={2026},
  pages={72468--72510},
  year={2026}
}

@article{cheng2024trace,
  title={Trace is the next autodiff: Generative optimization with rich feedback, execution traces, and llms},
  author={Cheng, Ching-An and Nie, Allen and Swaminathan, Adith},
  journal={Advances in Neural Information Processing Systems},
  volume={37},
  pages={71596--71642},
  year={2024}
}

@article{jones1998efficient,
  title={Efficient global optimization of expensive black-box functions},
  author={Jones, Donald R and Schonlau, Matthias and Welch, William J},
  journal={Journal of Global optimization},
  volume={13},
  number={4},
  pages={455--492},
  year={1998},
  publisher={Springer}
}

@article{snoek2012practical,
  title={Practical bayesian optimization of machine learning algorithms},
  author={Snoek, Jasper and Larochelle, Hugo and Adams, Ryan},
  journal={Advances in neural information processing systems},
  volume={25},
  year={2012}
}

@inproceedings{gardner2014bayesian,
  title={Bayesian optimization with inequality constraints.},
  author={Gardner, Jacob R and Kusner, Matt J and Xu, Zhixiang Eddie and Weinberger, Kilian Q and Cunningham, John P},
  booktitle={Icml},
  volume={2014},
  pages={937--945},
  year={2014}
}

@article{eriksson2019scalable,
  title={Scalable global optimization via local Bayesian optimization},
  author={Eriksson, David and Pearce, Michael and Gardner, Jacob and Turner, Ryan D and Poloczek, Matthias},
  journal={Advances in neural information processing systems},
  volume={32},
  year={2019}
}

@article{hase2018phoenics,
  title={Phoenics: a Bayesian optimizer for chemistry},
  author={Hase, Florian and Roch, Lo{\"\i}c M and Kreisbeck, Christoph and Aspuru-Guzik, Al{\'a}n},
  journal={ACS central science},
  volume={4},
  number={9},
  pages={1134--1145},
  year={2018},
  publisher={ACS Publications}
}

@article{shields2021bayesian,
  title={Bayesian reaction optimization as a tool for chemical synthesis},
  author={Shields, Benjamin J and Stevens, Jason and Li, Jun and Parasram, Marvin and Damani, Farhan and Alvarado, Jesus I Martinez and Janey, Jacob M and Adams, Ryan P and Doyle, Abigail G},
  journal={Nature},
  volume={590},
  number={7844},
  pages={89--96},
  year={2021},
  publisher={Nature Publishing Group UK London}
}

@article{granda2018controlling,
  title={Controlling an organic synthesis robot with machine learning to search for new reactivity},
  author={Granda, Jaros{\l}aw M and Donina, Liva and Dragone, Vincenza and Long, De-Liang and Cronin, Leroy},
  journal={Nature},
  volume={559},
  number={7714},
  pages={377--381},
  year={2018},
  publisher={Nature Publishing Group UK London}
}

@article{coley2019robotic,
  title={A robotic platform for flow synthesis of organic compounds informed by AI planning},
  author={Coley, Connor W and Thomas III, Dale A and Lummiss, Justin AM and Jaworski, Jonathan N and Breen, Christopher P and Schultz, Victor and Hart, Travis and Fishman, Joshua S and Rogers, Luke and Gao, Hanyu and others},
  journal={Science},
  volume={365},
  number={6453},
  pages={eaax1566},
  year={2019},
  publisher={American Association for the Advancement of Science}
}

@article{burger2020mobile,
  title={A mobile robotic chemist},
  author={Burger, Benjamin and Maffettone, Phillip M and Gusev, Vladimir V and Aitchison, Catherine M and Bai, Yang and Wang, Xiaoyan and Li, Xiaobo and Alston, Ben M and Li, Buyi and Clowes, Rob and others},
  journal={Nature},
  volume={583},
  number={7815},
  pages={237--241},
  year={2020},
  publisher={Nature Publishing Group UK London}
}

@article{macleod2020self,
  title={Self-driving laboratory for accelerated discovery of thin-film materials},
  author={MacLeod, Benjamin P and Parlane, Fraser GL and Morrissey, Thomas D and H{\"a}se, Florian and Roch, Lo{\"\i}c M and Dettelbach, Kevan E and Moreira, Raphaell and Yunker, Lars PE and Rooney, Michael B and Deeth, Joseph R and others},
  journal={Science Advances},
  volume={6},
  number={20},
  pages={eaaz8867},
  year={2020},
  publisher={American Association for the Advancement of Science}
}

@article{merchant2023scaling,
  title={Scaling deep learning for materials discovery},
  author={Merchant, Amil and Batzner, Simon and Schoenholz, Samuel S and Aykol, Muratahan and Cheon, Gowoon and Cubuk, Ekin Dogus},
  journal={Nature},
  volume={624},
  number={7990},
  pages={80--85},
  year={2023},
  publisher={Nature Publishing Group UK London}
}

@article{andrychowicz2016learning,
  title={Learning to learn by gradient descent by gradient descent},
  author={Andrychowicz, Marcin and Denil, Misha and Gomez, Sergio and Hoffman, Matthew W and Pfau, David and Schaul, Tom and Shillingford, Brendan and De Freitas, Nando},
  journal={Advances in neural information processing systems},
  volume={29},
  year={2016}
}

@article{chen2023symbolic,
  title={Symbolic discovery of optimization algorithms},
  author={Chen, Xiangning and Liang, Chen and Huang, Da and Real, Esteban and Wang, Kaiyuan and Pham, Hieu and Dong, Xuanyi and Luong, Thang and Hsieh, Cho-Jui and Lu, Yifeng and others},
  journal={Advances in neural information processing systems},
  volume={36},
  pages={49205--49233},
  year={2023}
}

@article{fawzi2022discovering,
  title={Discovering faster matrix multiplication algorithms with reinforcement learning},
  author={Fawzi, Alhussein and Balog, Matej and Huang, Aja and Hubert, Thomas and Romera-Paredes, Bernardino and Barekatain, Mohammadamin and Novikov, Alexander and R. Ruiz, Francisco J and Schrittwieser, Julian and Swirszcz, Grzegorz and others},
  journal={Nature},
  volume={610},
  number={7930},
  pages={47--53},
  year={2022},
  publisher={Nature Publishing Group UK London}
}

@article{mankowitz2023faster,
  title={Faster sorting algorithms discovered using deep reinforcement learning},
  author={Mankowitz, Daniel J and Michi, Andrea and Zhernov, Anton and Gelmi, Marco and Selvi, Marco and Paduraru, Cosmin and Leurent, Edouard and Iqbal, Shariq and Lespiau, Jean-Baptiste and Ahern, Alex and others},
  journal={Nature},
  volume={618},
  number={7964},
  pages={257--263},
  year={2023},
  publisher={Nature Publishing Group UK London}
}
\bibliographystyle{plainnat}

\def\FROMMAIN{}

\ifdefined\FROMMAIN
\else
    \documentclass[10pt]{article}

    \usepackage[margin=1in]{geometry}

    \usepackage{amsmath}
    \usepackage{amssymb}
    \usepackage{mathtools}
    \usepackage{bbm}

    \usepackage{booktabs}
    \usepackage{array}
    \usepackage{tabularx}
    \usepackage{multirow}
    \usepackage[table]{xcolor}
    \usepackage{graphicx}
    \usepackage{caption}

    \usepackage{algorithm}
    \usepackage{algorithmic}

    \usepackage{hyperref}

    \begin{document}
\fi


\clearpage
\appendix


\section{Scientific Objective and Simulator Specification}
\label{app:scientific_objective}

This appendix provides the full scientific formulation underlying the
gravitational-wave detector design task used throughout the paper.
We first define the finite-budget optimization problem, and then describe
the detector representation, continuous design variables, differentiable
physics simulator, native scientific objective, physical feasibility
constraints, and optimization coordinates.

\subsection{Finite-Budget Scientific Optimization}
\label{app:finite_budget_objective}

For a detector topology $\tau \in \mathcal{T}$, let
$x \in \mathcal{X}_{\tau}$ denote its continuous design variables,
$L_{\tau}(x)$ the native scientific objective, and
$\mathcal{F}_{\tau} \subseteq \mathcal{X}_{\tau}$ the physically feasible
subset. The unconstrained scientific design problem can be written as
\begin{equation}
    x_{\tau}^{\star}
    =
    \arg\min_{x \in \mathcal{F}_{\tau}}
    L_{\tau}(x).
    \label{eq:supp_native_objective}
\end{equation}

In practice, however, scientific optimization is performed under a finite
evaluation budget. Performance therefore depends not only on the local
parameter updates, but also on where the search starts and which optimization
procedure is used. We represent the complete search policy as
\begin{equation}
    \pi = (W,O),
    \label{eq:supp_policy}
\end{equation}
where $W$ denotes the selection policy that determines the initialization
$x_0$, and $O$ denotes the optimizer that subsequently generates a trajectory
\begin{equation}
    \mathcal{T}_{\pi}
    =
    \{x_0,x_1,\ldots,x_T\}.
\end{equation}

Let $c_t$ denote the cumulative scientific evaluation cost when $x_t$ has
been evaluated, and let $B$ denote the total evaluation budget. We define the
finite-budget performance of policy $\pi$ on target $\tau$ as
\begin{equation}
    J_{\tau}(\pi;B)
    =
    \min_{\substack{
        t:\,c_t\le B\\
        x_t\in\mathcal{F}_{\tau}
    }}
    L_{\tau}(x_t),
    \qquad
    \min \varnothing := +\infty .
    \label{eq:supp_budget_objective}
\end{equation}

Equation~\eqref{eq:supp_budget_objective} evaluates the best physically
feasible design obtained before the budget is exhausted. For a fixed execution
trace and $B_2 \ge B_1$, the feasible incumbent set available at $B_1$ is a
subset of that available at $B_2$, and therefore
\begin{equation}
    J_{\tau}(\pi;B_2)
    \le
    J_{\tau}(\pi;B_1).
\end{equation}
This monotonicity applies to the recorded best feasible solution, rather than
to every intermediate optimization iterate.

ADAPT optimizes the two components of the search policy jointly:
\begin{equation}
    \left(
        W_{\tau}^{\star},
        O_{\tau}^{\star}
    \right)
    \in
    \arg\min_{
        W\in\mathcal{W},
        \,
        O\in\mathcal{O}
    }
    J_{\tau}\bigl((W,O);B\bigr).
    \label{eq:supp_policy_objective}
\end{equation}
The role of $W_{\tau}^{\star}$ is to determine \emph{where to search}, whereas
$O_{\tau}^{\star}$ determines \emph{how to search}. Importantly,
Equation~\eqref{eq:supp_policy_objective} defines the objective of the
meta-level search; it does not assume that the discrete policy space is
differentiable or that the global optimum over $\mathcal{W}\times\mathcal{O}$
can be found.

\subsection{Detector Topology Representation}
\label{app:topology_representation}

A detector target consists of a discrete optical topology together with a set
of continuous design variables. In the size-three UIFO representation used in
this work, each topology is encoded by a string containing nine internal
component symbols followed by twelve boundary symbols.

The nine internal positions follow the order
\begin{equation}
    11,\ 12,\ 13,\ 21,\ 22,\ 23,\ 31,\ 32,\ 33 ,
\end{equation}
and the twelve boundary positions follow
\begin{equation}
    01,\ 02,\ 03,\ 10,\ 14,\ 20,\ 24,\ 30,\ 34,\ 41,\ 42,\ 43 .
\end{equation}

For internal components, symbols A--D represent beam splitters with different
orientations, whereas E--H represent directional beam splitters with the
corresponding orientation convention. Boundary symbols specify optical
sources or readout elements:
\begin{equation}
    \mathrm{L}=\text{laser},\qquad
    \mathrm{S}=\text{squeezer},\qquad
    \mathrm{D}=\text{detector},\qquad
    \mathrm{H}=\text{balanced homodyne readout}.
\end{equation}

For example, the eight held-out targets used in the main experiments are
distinct discrete topologies, while optimization is performed over their
topology-specific continuous parameter spaces. Complete visualizations of all
eight targets, including the initial configuration, the standard Adam result,
and the final ADAPT result, are provided in Appendix~\ref{app:topology_atlas}.

\subsection{Continuous Design Variables and Physical Bounds}
\label{app:parameter_bounds}

For a fixed topology $\tau$, the continuous design vector
$x\in\mathbb{R}^{d_{\tau}}$ contains the active optical and mechanical
parameters associated with that topology. Depending on the included optical
components, these variables may include mirror reflectivities and tunings,
squeezing parameters, laser powers, test-mass masses, and propagation lengths.

Table~\ref{tab:supp_parameter_bounds} summarizes the default physical parameter
ranges used by the simulator.

\begin{table}[t]
    \centering
    \caption{
    \textbf{Continuous detector-design variables and default physical bounds.}
    The active parameter dimension $d_{\tau}$ depends on the detector topology.
    Some physical properties may share one optimization scalar and are therefore
    optimized jointly.
    }
    \label{tab:supp_parameter_bounds}
    \small
    \setlength{\tabcolsep}{7pt}
    \renewcommand{\arraystretch}{1.15}
    \begin{tabular}{lll}
        \toprule
        \textbf{Parameter} &
        \textbf{Range} &
        \textbf{Unit} \\
        \midrule
        Reflectivity
            & $[10^{-12},\,1-10^{-12}]$
            & dimensionless \\
        Mirror tuning
            & $[-360,\,360]$
            & degree \\
        Squeezing level
            & $[0,\,10]$
            & dB \\
        Squeezing angle
            & $[-360,\,360]$
            & degree \\
        Laser power
            & $[0,\,200]$
            & W \\
        Mass
            & $[0.01,\,200]$
            & kg \\
        Propagation length
            & $[0.1,\,4000]$
            & m \\
        \bottomrule
    \end{tabular}
\end{table}

The nominal physical range of a reflectivity is $[0,1]$. In the numerical
implementation, both endpoints are moved inward by $10^{-12}$ to avoid
singular boundary behavior.

The parameter dimension is topology-dependent. In addition, several optical
properties can be coupled to the same optimization scalar. Consequently,
$d_{\tau}$ is determined from the topology-specific optimization schema rather
than being fixed globally. When historical designs are transferred between
topologies, these coupled parameter groups are preserved; the corresponding
procedure is described in Appendix~\ref{app:semantic_transfer}.

\subsection{Differentiable Physics Simulator}
\label{app:physics_simulator}

Each candidate detector is evaluated using the differentiable interferometer
simulator employed throughout the study. For every design $x$, the simulator
constructs quantum-noise, amplitude-modulation, and frequency-modulation
configurations and evaluates their corresponding detector responses over the
same frequency grid.

We use $N_f=100$ logarithmically spaced frequencies from
$f_{\min}=20~\mathrm{Hz}$ to $f_{\max}=5000~\mathrm{Hz}$:
\begin{equation}
    f_k
    =
    f_{\min}
    \left(
        \frac{f_{\max}}{f_{\min}}
    \right)^{\frac{k-1}{N_f-1}},
    \qquad
    k=1,\ldots,N_f .
    \label{eq:supp_frequency_grid}
\end{equation}

Let $N_q(f;x)$ denote the simulated quantum-noise magnitude, and let
$D_q(f;x)$, $D_a(f;x)$, and $D_f(f;x)$ denote the detector transfer magnitudes
extracted from the quantum, amplitude-modulation, and frequency-modulation
simulations, respectively. The resulting strain sensitivity is
\begin{equation}
    H(f;x)
    =
    \frac{
        \sqrt{
            |N_q(f;x)|^2
            +
            \left(4\times10^{-9}D_a(f;x)\right)^2
            +
            \left(10^{-8}fD_f(f;x)\right)^2
        }
    }{
        D_q(f;x)
    }.
    \label{eq:supp_sensitivity}
\end{equation}

For numerical stability, the extracted detector transfer is lower bounded by
a small positive constant before division. The coefficients in
Equation~\eqref{eq:supp_sensitivity} are fixed simulator conventions and are
not optimized or learned by ADAPT.

Because the simulator is differentiable with respect to the continuous design
variables, gradient-based optimizers can directly differentiate the scientific
objective through the complete physical simulation.

\subsection{Native Sensitivity Objective}
\label{app:native_loss}

Let $H_{\mathrm{ref}}(f)$ denote the fixed reference sensitivity curve.
The native sensitivity objective is the mean logarithmic sensitivity ratio
across the $N_f$ evaluation frequencies:
\begin{equation}
    L_{\mathrm{sens}}(x)
    =
    \frac{1}{N_f}
    \sum_{k=1}^{N_f}
    \log_{10}
    \frac{
        H(f_k;x)
    }{
        H_{\mathrm{ref}}(f_k)
    }.
    \label{eq:supp_sensitivity_loss}
\end{equation}

Equation~\eqref{eq:supp_sensitivity_loss} can equivalently be written as
\begin{equation}
    L_{\mathrm{sens}}(x)
    =
    \log_{10}
    \left[
        \prod_{k=1}^{N_f}
        \frac{
            H(f_k;x)
        }{
            H_{\mathrm{ref}}(f_k)
        }
    \right]^{1/N_f}.
    \label{eq:supp_geometric_interpretation}
\end{equation}
Thus,
\begin{equation}
    10^{L_{\mathrm{sens}}(x)}
\end{equation}
is the geometric mean of the sensitivity ratio relative to the reference
detector. This objective therefore rewards broadband improvement rather than
performance at only a single frequency.

Away from nonsmooth numerical boundaries, its gradient satisfies
\begin{equation}
    \nabla_x L_{\mathrm{sens}}(x)
    =
    \frac{1}{N_f\ln 10}
    \sum_{k=1}^{N_f}
    \frac{
        \nabla_x H(f_k;x)
    }{
        H(f_k;x)
    }.
    \label{eq:supp_sensitivity_gradient}
\end{equation}
The reference curve contributes no gradient because it is fixed throughout
optimization.

\subsection{Physical Constraints and Power Penalty}
\label{app:physical_constraints}

In addition to sensitivity, candidate designs must satisfy internal optical
power constraints. Let $P_{g,j}(x)$ denote the power at constrained port $j$
belonging to constraint group $g$, and let $T_g$ be its corresponding physical
threshold.

The simulator uses the following threshold levels:
\begin{equation}
    T_{\mathrm{hard}}
    =
    3.5\times10^6~\mathrm{W},
    \qquad
    T_{\mathrm{soft}}
    =
    2\times10^3~\mathrm{W},
    \qquad
    T_{\mathrm{detector}}
    =
    10^{-2}~\mathrm{W}.
    \label{eq:supp_power_thresholds}
\end{equation}

For each constrained power, define the normalized violation
\begin{equation}
    a_{g,j}(x)
    =
    \max
    \left(
        \frac{P_{g,j}(x)}{T_g}-1,
        0
    \right),
\end{equation}
and the bounded penalty
\begin{equation}
    p_{g,j}(x)
    =
    \frac{
        a_{g,j}(x)
    }{
        1+a_{g,j}(x)
    }.
    \label{eq:supp_power_penalty}
\end{equation}

The complete native optimization objective is
\begin{equation}
    L_{\tau}(x)
    =
    L_{\mathrm{sens}}(x)
    +
    \sum_{g,j}
    p_{g,j}(x).
    \label{eq:supp_total_loss}
\end{equation}

Each individual penalty satisfies $0\le p_{g,j}(x)<1$, while the complete
penalty can exceed one because multiple physical constraints may be violated
simultaneously.

For $P_{g,j}>T_g$, Equation~\eqref{eq:supp_power_penalty} reduces to
\begin{equation}
    p_{g,j}
    =
    1-\frac{T_g}{P_{g,j}},
\end{equation}
and therefore
\begin{equation}
    \frac{\partial p_{g,j}}{\partial P_{g,j}}
    =
    \frac{T_g}{P_{g,j}^{2}}.
    \label{eq:supp_penalty_gradient}
\end{equation}
Below the threshold, the derivative is zero.

The penalty in Equation~\eqref{eq:supp_total_loss} facilitates differentiable
optimization, but physical feasibility is evaluated independently. A design is
considered feasible only if
\begin{equation}
    x\in\mathcal{F}_{\tau}
    \quad\Longleftrightarrow\quad
    x\in\mathcal{X}_{\tau}
    \ \text{and}\
    P_{g,j}(x)\le T_g
    \quad
    \forall(g,j).
    \label{eq:supp_feasibility}
\end{equation}

This distinction is important: the differentiable penalty guides search,
whereas the final scientific validity of a design is determined by the
physical constraints themselves.

\subsection{Bounded-to-Unbounded Optimization Coordinates}
\label{app:coordinate_transform}

The physical parameters lie inside heterogeneous bounded intervals. To provide
a common unconstrained optimization space, we transform each physical parameter
$x_j\in[\ell_j,h_j]$ to an unconstrained coordinate $u_j$.

First define
\begin{equation}
    q_j
    =
    \operatorname{clip}
    \left(
        \frac{x_j-\ell_j}{h_j-\ell_j},
        \epsilon_u,
        1-\epsilon_u
    \right),
\end{equation}
where $\epsilon_u=10^{-9}$ is used for numerical stability. The unconstrained
coordinate is then
\begin{equation}
    u_j
    =
    \log q_j-\log(1-q_j).
    \label{eq:supp_logit_transform}
\end{equation}
The inverse transformation is
\begin{equation}
    x_j(u_j)
    =
    \ell_j
    +
    (h_j-\ell_j)\sigma(u_j),
    \qquad
    \sigma(u)
    =
    \frac{1}{1+e^{-u}}.
    \label{eq:supp_inverse_transform}
\end{equation}

By the chain rule,
\begin{equation}
    \nabla_u L
    =
    \operatorname{diag}
    \left[
        (h-\ell)
        \odot
        \sigma(u)
        \odot
        \bigl(1-\sigma(u)\bigr)
    \right]
    \nabla_x L.
    \label{eq:supp_coordinate_gradient}
\end{equation}

If one scalar controls multiple coupled optical properties
$\{y_k\}$, their gradient contributions are accumulated:
\begin{equation}
    \frac{\partial L}{\partial x_j}
    =
    \sum_k
    \frac{\partial L}{\partial y_k}
    \frac{\partial y_k}{\partial x_j}.
    \label{eq:supp_coupled_gradient}
\end{equation}
For this reason, coupled parameter groups are treated as indivisible semantic
units when transferring historical designs between detector topologies.

\section{Historical Experience Library and Target-Aware Retrieval}
\label{app:historical_library}

ADAPT does not initialize every scientific search from scratch. Instead, it
constructs a target-specific prior from a historical library of previously
evaluated detector designs. This appendix describes the composition of the
library, target-aware retrieval, topology-compatible parameter transfer,
prior-based completion of unmatched parameters, and the construction of the
candidate initialization pool used by the Selection Evolver.

\subsection{Historical Design Library}
\label{app:library_composition}

The historical experience library used in this work contains
\begin{equation}
    15{,}409
\end{equation}
transferable gravitational-wave detector designs.

The library is formed from two sources:
\begin{equation}
    9{,}342
    \quad\text{official training representatives},
\end{equation}
and
\begin{equation}
    6{,}067
    \quad\text{compatible historical GraviTune designs}.
\end{equation}
Together,
\begin{equation}
    9{,}342 + 6{,}067 = 15{,}409.
\end{equation}

The complete retrieval index contains $16{,}727$ historical topology names.
Among the additional GraviTune entries, $1{,}318$ are incompatible with the
parameter-transfer schema and therefore cannot provide valid transferred
initializations. They are excluded when constructing the usable historical
design bank.

For reproducibility, the compressed historical bank used by the current
release has SHA-256
\begin{quote}
\scriptsize
\texttt{
6959bfcb3f24e1b963e3e0b3dc4f0ada414fca0dc9b9426f6cfd566e4f624594
}.
\end{quote}

The historical library is used exclusively as a source of prior scientific
experience. Its size should therefore be interpreted as the number of reusable
scientific designs, not as a language-model training set or a number of LLM
training examples.

\subsection{Target-Aware Topology Retrieval}
\label{app:target_retrieval}

For a new target topology $\tau$, the Target-Aware Matcher first ranks
historical designs according to structural similarity and retrieves the
$K=64$ closest compatible designs. These designs form the candidate pool
\begin{equation}
    \mathcal{C}_{\tau}
    =
    \left\{
        d_{(1)},d_{(2)},\ldots,d_{(64)}
    \right\}.
    \label{eq:supp_candidate_pool}
\end{equation}


In the topology-aware retrieval implementation, the distance decomposes into
the nine internal components and twelve boundary components. Let
$c_{\tau,j}$ and $b_{\tau,j}$ denote the internal and boundary symbols of the
target, respectively, and let $c_{i,j}$ and $b_{i,j}$ denote those of a
historical topology $i$. The retrieval distance is
\begin{equation}
    D(\tau,i)
    =
    \frac{1}{9}
    \sum_{j=1}^{9}
    d_c(c_{\tau,j},c_{i,j})
    +
    \frac{1}{12}
    \sum_{j=1}^{12}
    \mathbbm{1}
    \left[
        b_{\tau,j}\neq b_{i,j}
    \right],
    \label{eq:supp_topology_distance}
\end{equation}
where
\begin{equation}
    d_c(a,b)
    =
    \begin{cases}
        0,
        & a=b,\\[2mm]
        0.25,
        & a\neq b
          \text{ but }a\text{ and }b
          \text{ belong to the same component family},\\[2mm]
        1,
        & \text{otherwise}.
    \end{cases}
    \label{eq:supp_internal_distance}
\end{equation}

This definition distinguishes an orientation change within the same optical
component family from a replacement by a different component family.
Exact matches therefore receive zero cost, orientation-level changes receive
a smaller cost, and changes of component type receive the largest cost.
Symbols are compared at fixed positions without rotational or reflection
alignment.

Historical candidates are ranked by increasing
$D(\tau,i)$. Exact copies of the target topology are excluded from retrieval,
and incompatible historical records are skipped. Retrieval continues until
64 distinct transferable sources have been collected, without dataset-specific
quotas. Lexicographic ordering of topology strings is used to resolve
equal-distance ties.

\subsection{Topology-Compatible Semantic Transfer}
\label{app:semantic_transfer}

A retrieved historical detector cannot in general be copied directly to a new
topology because the target may contain a different set of optical components
and a different number of continuous parameters. We therefore transfer
historical designs according to the semantic identity of each optimization
variable.

Each scalar is associated with
\begin{equation}
    \bigl(
        \text{physical property},
        \text{component names and types},\,
        \text{complete controlled component set}
    \bigr),
\end{equation}
where the controlled component set contains the complete group of optical
properties coupled to that scalar. Two variables are considered directly
transferable only when their semantic identities match and the historical
value is valid under the target bounds.

For target coordinate $j$ and retrieved historical design $i$, the transferred
value is
\begin{equation}
    x_{\tau,j}^{(i)}
    =
    \begin{cases}
        x_{i,k},
        &
        \text{if a compatible semantic match }k\text{ exists},
        \\[2mm]
        \ell_{\tau,j}
        +
        \left(
            h_{\tau,j}-\ell_{\tau,j}
        \right)m_j,
        &
        \text{otherwise},
    \end{cases}
    \label{eq:supp_semantic_transfer}
\end{equation}
where $m_j\in[0,1]$ is a normalized historical prior used to complete unmatched
coordinates.

Preserving the complete coupled component set is essential. If one scalar
jointly controls several optical properties, transferring only one member of
the group would alter the physical meaning of the optimization variable and
would no longer reproduce a valid target parameterization.

\subsection{Hierarchical Prior for Unmatched Parameters}
\label{app:hierarchical_prior}

Not every target coordinate has an exact counterpart in every retrieved
historical topology. Instead of filling such coordinates with an arbitrary
constant or random value, ADAPT uses frozen training-set normalized medians.
All prior statistics are computed from the training set only.

For each physical parameter, historical values are first normalized to
$[0,1]$ using their corresponding physical bounds. The completion prior is
selected hierarchically:

\begin{enumerate}
    \item \textbf{P2: component-type/property prior.}
    Use the median normalized value for the same component type and physical
    property whenever this statistic is available.

    \item \textbf{P1: property-level prior.}
    If a component-type-specific statistic is unavailable, use the median
    normalized value for the same physical property across compatible
    components.

    \item \textbf{P0: global prior.}
    If neither P2 nor P1 is available, use the global normalized median.
\end{enumerate}

For coupled variables spanning multiple component types, the component-specific
P2 prior is not used because it would assign a single coupled scalar according
to only one member of the group. In this case, the hierarchy falls back to the
property-level or global prior.

This procedure allows a historical detector to provide a complete
topology-compatible initialization even when the source and target topologies
are not identical.

\subsection{Target-Specific Screening of Retrieved Designs}
\label{app:target_screening}

Retrieval similarity alone does not imply that a transferred design performs
well on the new target. The 64 transferred candidates are therefore evaluated
once under the \emph{target} simulator without optimizer updates before the
Selection Evolver allocates the remaining optimization budget. This screening
is charged to the same total wall-clock budget as subsequent optimization.

For candidate $i$, let $L_i$ denote its target loss and define its maximum
normalized physical violation as
\begin{equation}
    v_i
    =
    \max
    \left(
        0,\,
        \max_{g,j}
        \left[
            \frac{P_{g,j}(x_i)}{T_g}-1
        \right]
    \right).
    \label{eq:supp_violation_score}
\end{equation}

Candidates are first separated according to physical feasibility. A finite
feasible candidate is preferred to an infeasible candidate, and an infeasible
candidate is preferred to an invalid evaluation. The ranking can be written
lexicographically as
\begin{equation}
    k_i
    =
    \begin{cases}
        (0,L_i,0,i),
        & \text{if }x_i\in\mathcal{F}_{\tau},\\[1mm]
        (1,v_i,L_i,i),
        & \text{if the evaluation is finite but infeasible},\\[1mm]
        (2,+\infty,+\infty,i),
        & \text{if the evaluation is invalid}.
    \end{cases}
    \label{eq:supp_screening_key}
\end{equation}

This target-side screening step is important because historical performance is
used only as prior experience; it is not assumed to remain valid after
transferring a design to a different detector topology.

\subsection{Quality--Diversity Construction of the Eight-Start Set}
\label{app:quality_diversity}

When at least eight feasible candidates are available after screening, ADAPT
constructs an eight-start set consisting of four quality-oriented candidates
and four diversity-oriented candidates.

The four quality candidates are the feasible transferred designs with the
lowest target loss:
\begin{equation}
    \mathcal{S}_{\mathrm{quality}}
    =
    \operatorname{Top4}_{x_i\in\mathcal{F}_{\tau}}
    \left[
        -L_i
    \right].
\end{equation}

To avoid concentrating all starts in the same local basin, four additional
candidates are selected greedily according to their distance from the current
selected set. Let $q(x_i)\in[0,1]^{d_{\tau}}$ denote the target design after
normalization by its physical parameter bounds. We use the root-mean-square
parameter distance
\begin{equation}
    d_q(i,j)
    =
    \sqrt{
        \frac{1}{d_{\tau}}
        \sum_{k=1}^{d_{\tau}}
        \left(
            q_k(x_i)-q_k(x_j)
        \right)^2
    }.
    \label{eq:supp_parameter_distance}
\end{equation}

Starting with
$\mathcal{S}=\mathcal{S}_{\mathrm{quality}}$, each diversity candidate is
selected as
\begin{equation}
    i^{\star}
    =
    \arg\max_{i:\,x_i\in\mathcal{F}_{\tau},\,i\notin\mathcal{S}}
    \min_{j\in\mathcal{S}}
    d_q(i,j),
    \label{eq:supp_farthest_point}
\end{equation}
and is then appended to $\mathcal{S}$. This operation is repeated until
$|\mathcal{S}|=8$. Ties are resolved by candidate index.

The resulting initialization set can therefore be written as
\begin{equation}
    \mathcal{S}_{8}
    =
    \underbrace{
        \mathcal{S}_{\mathrm{quality}}
    }_{\text{4 low-loss starts}}
    \cup
    \underbrace{
        \mathcal{S}_{\mathrm{diversity}}
    }_{\text{4 coverage-preserving starts}} .
    \label{eq:supp_four_plus_four}
\end{equation}

If fewer than eight feasible candidates are available, all feasible candidates
are retained first and the remaining positions are filled with numerically
valid infeasible candidates in order of physical violation, target loss, and
candidate index. In this fallback case, the four-plus-four decomposition is
not artificially enforced. Fewer than eight numerically valid candidates
terminates the selection run.

The eight selected starts are passed to the Selection Evolver, which performs
the subsequent progressive
\begin{equation}
    8\rightarrow2\rightarrow1
\end{equation}
allocation described in Appendix~\ref{app:selection_evolution}.

\subsection{Algorithmic Summary}
\label{app:retrieval_algorithm}

The complete procedure from a new detector topology to the eight-start
initialization set is summarized in Algorithm~\ref{alg:supp_retrieval}.

\begin{algorithm}[t]
\caption{Target-aware retrieval and initialization construction}
\label{alg:supp_retrieval}
\begin{algorithmic}[1]
\REQUIRE Target topology $\tau$, historical library $\mathcal{D}$,
         target parameter schema, physical bounds
\ENSURE Eight target-specific candidate starts $\mathcal{S}_8$

\STATE Rank historical topologies by target-aware topology distance,
       breaking ties lexicographically
\STATE Exclude the exact target and incompatible historical records
\STATE Retrieve the first $64$ transferable historical designs

\FOR{each retrieved design $d_i$}
    \STATE Transfer semantically matched target parameters
    \STATE Fill unmatched parameters using the P2$\rightarrow$P1$\rightarrow$P0 prior
    \STATE Evaluate the transferred design on the target simulator
    \STATE Record target loss, feasibility, and physical-violation statistics
\ENDFOR

\IF{fewer than eight transferred candidates are numerically valid}
    \STATE Terminate the selection run
\ENDIF

\IF{at least eight transferred candidates are feasible}
    \STATE Select the four lowest-loss feasible candidates
    \STATE Add four candidates by greedy farthest-point selection
\ELSE
    \STATE Retain all feasible candidates
    \STATE Fill remaining slots with finite infeasible candidates by increasing
           violation, target loss, and candidate index
\ENDIF

\STATE \textbf{return} $\mathcal{S}_8$
\end{algorithmic}
\end{algorithm}

\subsection{Isolation of Held-Out Targets}
\label{app:heldout_isolation}

The historical library provides prior scientific experience, but held-out
outcomes are not used as historical supervision. For a held-out target
$\tau_{\mathrm{test}}$, its topology is available at inference time because it
defines the scientific problem to be optimized. However, its held-out optimized
design, final loss, future optimization trajectory, and validator outcome are
not included in the historical source library.

Accordingly, target-aware retrieval uses only the target topology as a query:
\begin{equation}
    \tau_{\mathrm{test}}
    \longrightarrow
    \mathcal{C}_{\tau_{\mathrm{test}}},
\end{equation}
while the retrieved records themselves originate from the permitted historical
experience bank.

The same principle applies to all subsequent validation. Information obtained
from the final held-out validator is used only for reporting the scientific
performance of the resulting policy and is not fed back into the historical
library or used to revise the reported policy.


\section{Selection Evolution: Learning Where to Search}
\label{app:selection_evolution}

The first adaptive stage of ADAPT determines \emph{where to search}.
Rather than treating initialization as a fixed preprocessing choice,
the Selection Evolver searches over policies that transform historical
scientific experience into a target-specific starting point under a finite
evaluation budget.

The complete stage contains two conceptually distinct levels.
During method development, agentic workflows propose and refine candidate
selection policies using development targets and execution feedback.
After development, the resulting selection policy is frozen.
At held-out evaluation time, this fixed policy is applied to the
target-specific candidate pool constructed in
Appendix~\ref{app:historical_library}.
No held-out outcome is used to revise the selection policy itself.

For a target $\tau$, the Selection Evolver therefore implements the mapping
\begin{equation}
    W^{\star}:
    \mathcal{C}_{\tau}
    \longrightarrow
    x_{0,\tau}^{\star},
    \label{eq:supp_selection_mapping}
\end{equation}
where $\mathcal{C}_{\tau}$ contains the 64 retrieved historical designs and
$x_{0,\tau}^{\star}$ is the parameter vector handed to the subsequent
Optimizer Evolver.

\subsection{Selection Policy as a Search Object}
\label{app:selection_policy_space}

A conventional optimization pipeline typically fixes the initialization
strategy before optimization begins. In ADAPT, the initialization strategy
itself is treated as an object of meta-level search.

We denote a candidate selection policy at evolution step $k$ by
\begin{equation}
    W^{(k)}.
\end{equation}
A selection policy may specify how historical candidates are filtered,
how quality and diversity are combined, how short optimization probes are
allocated, how weak branches are pruned, and how exploratory branches are
protected.

Executing $W^{(k)}$ produces an empirical feedback record
\begin{equation}
    F_W^{(k)}
    =
    \mathcal{E}
    \left(
        W^{(k)}
    \right),
    \label{eq:supp_selection_feedback}
\end{equation}
where $\mathcal{E}$ denotes execution against the differentiable scientific
simulator under the assigned development budget.

The feedback contains measured scientific outcomes rather than a language-model
preference. At minimum, the execution trace records objective values,
physical feasibility, optimization progress, consumed evaluation budget,
and failure status. The proposing workflow receives a compact representation
of these observations and uses them to revise the policy:
\begin{equation}
    W^{(k)}
    \xrightarrow{\;\mathcal{E}\;}
    F_W^{(k)}
    \xrightarrow{\;\text{agentic revision}\;}
    W^{(k+1)}.
    \label{eq:supp_selection_evolution_loop}
\end{equation}

Strong previously evaluated policies remain available for comparison with
new proposals. Policy evolution is therefore driven by measured scientific
performance and does not rely on an LLM self-assessment of the generated
policy.

\subsection{Agentic Workflow Pool}
\label{app:selection_agent_pool}

During development, ADAPT considers four agentic workflow families:
\begin{equation}
    \mathcal{A}
    =
    \{
        \mathrm{GEPA},
        \mathrm{OpenEvolve},
        \mathrm{ShinkaEvolve},
        \mathrm{ASI}
    \}.
\end{equation}

These workflows are crossed with two language-model backends,
\begin{equation}
    \mathcal{M}
    =
    \{
        \mathrm{GPT\text{-}6\ Astra},
        \mathrm{Claude\text{-}fable\text{-}5.1}
    \},
\end{equation}
and two fixed reference optimizer configurations used to evaluate candidate
selection rules.

The resulting development pool therefore contains
\begin{equation}
    |\mathcal{A}|
    \times
    |\mathcal{M}|
    \times
    2
    =
    4\times2\times2
    =
    16
\end{equation}
workflow configurations.

The role of these configurations is to search over the \emph{selection
procedure}, rather than over the detector parameters directly.
The final held-out evaluation does not re-run this meta-level comparison.
Instead, the selection policy retained during development is frozen and
executed as part of the final ADAPT pipeline.

The complete system prompts and feedback templates used for selection-policy
evolution are reproduced in Appendix~\ref{app:prompts}.
Exact model and execution settings are summarized in
Appendix~\ref{app:experimental_setup}.

\subsection{From 64 Retrieved Designs to Eight Search Starts}
\label{app:selection_64_to_8}

For a target $\tau$, the Target-Aware Matcher first constructs the retrieved
candidate pool
\begin{equation}
    \mathcal{C}_{\tau}
    =
    \{x_1,\ldots,x_{64}\}.
\end{equation}

As described in Appendix~\ref{app:quality_diversity}, these transferred
candidates are evaluated on the target simulator and reduced to eight starting
points:
\begin{equation}
    \mathcal{S}_{8}
    =
    \mathcal{S}_{\mathrm{quality}}
    \cup
    \mathcal{S}_{\mathrm{diversity}},
    \qquad
    |\mathcal{S}_{8}|=8.
    \label{eq:supp_s8}
\end{equation}

When at least eight feasible transferred candidates are available,
$\mathcal{S}_{\mathrm{quality}}$ contains the four feasible candidates with
the lowest target loss and
$\mathcal{S}_{\mathrm{diversity}}$ contains four additional candidates
selected by the greedy farthest-point criterion in
Equation~\eqref{eq:supp_farthest_point}.

It is important to distinguish this step from the subsequent optimization
probes. The 64 historical records form a retrieval and screening pool;
ADAPT does not allocate a full optimization trajectory to all 64 candidates.
Only the resulting eight starts enter the progressive optimization schedule.

\subsection{Progressive $8\rightarrow2\rightarrow1$ Allocation}
\label{app:progressive_allocation}

The selected eight starts are evaluated through a progressive
budget-allocation policy:
\begin{equation}
    64
    \;\longrightarrow\;
    8
    \;\longrightarrow\;
    2
    \;\longrightarrow\;
    1.
    \label{eq:supp_64_8_2_1}
\end{equation}

Let the eight historical starts be
\begin{equation}
    \mathcal{H}_8
    =
    \{
        h_1,\ldots,h_8
    \}.
\end{equation}

Each historical start is initially optimized for
\begin{equation}
    T_{\mathrm{screen}}
    =
    180~\mathrm{s}
    =
    3~\mathrm{min}
    \label{eq:supp_screen_time}
\end{equation}
of active trajectory time using the same fixed screening optimizer and
scientific objective.

For branch $i$, define its incumbent feasible objective after active time $t$
as
\begin{equation}
    J_i(t)
    =
    \min_{
        \substack{
            s\le t\\
            x_{i,s}\in\mathcal{F}_{\tau}
        }
    }
    L_{\tau}(x_{i,s}),
    \qquad
    \min\varnothing:=+\infty.
    \label{eq:supp_branch_incumbent}
\end{equation}

After the initial screening stage, the two strongest historical trajectories
are retained:
\begin{equation}
    \mathcal{H}_2
    =
    \operatorname{Top2}_{h_i\in\mathcal{H}_8}
    \left[
        -J_i(T_{\mathrm{screen}})
    \right].
    \label{eq:supp_top2}
\end{equation}

If a branch has not produced a feasible design, finite infeasible candidates
are ordered using their physical-violation magnitude before invalid evaluations,
following the same feasibility-aware convention used in
Appendix~\ref{app:target_screening}.

Each of the two retained historical trajectories receives an additional
\begin{equation}
    T_{\mathrm{continue}}
    =
    1080~\mathrm{s}
    =
    18~\mathrm{min}
\end{equation}
of active optimization. Consequently, a retained historical trajectory may
receive up to
\begin{equation}
    T_{\mathrm{hist}}
    =
    T_{\mathrm{screen}}
    +
    T_{\mathrm{continue}}
    =
    1260~\mathrm{s}
    =
    21~\mathrm{min}
    \label{eq:supp_21min}
\end{equation}
before the final historical start is selected.

This distinction is important: the ``21-minute'' continuation reported in
the main paper denotes the cumulative \emph{active optimization time of one
surviving trajectory}, rather than 21 minutes of elapsed time from the start
of the complete ADAPT run.

\subsection{Protected Exploratory Branch}
\label{app:protected_exploration}

Aggressive pruning can prematurely concentrate the search around historical
designs that appear strong during a short screening interval.
To preserve an independent alternative, the final selection policy maintains
a protected exploratory branch in parallel with the historical candidates.

In the implemented scheduler, this branch is initialized independently of the
eight retrieved historical starts and receives a protected active-time quota
\begin{equation}
    T_{\mathrm{explore}}
    =
    3600~\mathrm{s}
    =
    60~\mathrm{min}.
    \label{eq:supp_exploration_budget}
\end{equation}

The exploratory trajectory is not removed merely because its early objective
is worse than those of the historical branches. Its compute is interleaved
with the retained historical trajectories in
\begin{equation}
    360~\mathrm{s}
    =
    6~\mathrm{min}
\end{equation}
scheduling slices.

The purpose of this branch is not to introduce a different scientific
objective. It is evaluated with the same simulator, physical constraints,
and native loss as every other branch. Its role is purely algorithmic:
it prevents the progressive scheduler from allocating the complete
pre-optimization budget exclusively according to early historical performance.

The selection mechanism therefore combines two complementary forms of
exploration:
\begin{equation}
    \underbrace{
        \mathcal{S}_{\mathrm{diversity}}
    }_{\text{diversity among historical designs}}
    \qquad\text{and}\qquad
    \underbrace{
        h_{\mathrm{explore}}
    }_{\text{independent protected trajectory}}.
\end{equation}

\subsection{Active-Time Accounting and the Shared Deadline}
\label{app:selection_time_accounting}

The branch quotas above are \emph{active optimization times}. They should not
be added directly to elapsed wall-clock time without specifying the scheduler.

Ignoring simulator and scheduling overhead, the nominal active compute allocated
before the final optimizer stage is
\begin{align}
    B_{\mathrm{selection}}
    &=
    8T_{\mathrm{screen}}
    +
    2T_{\mathrm{continue}}
    +
    T_{\mathrm{explore}}
    \nonumber\\
    &=
    8(180)
    +
    2(1080)
    +
    3600
    \nonumber\\
    &=
    7200~\mathrm{s}
    =
    120~\mathrm{min}.
    \label{eq:supp_selection_active_cost}
\end{align}

Equation~\eqref{eq:supp_selection_active_cost} is an accounting identity for
the scheduled active quotas; it is not an independent experimental result.

The complete pipeline is governed by a shared wall-clock deadline. Therefore,
candidate evaluation, simulator execution, synchronization, compilation when
charged to the run, bookkeeping, branch switching, selection, and the final
optimizer all consume the same end-to-end budget.

Let $T_{\mathrm{global}}$ denote the global deadline and
$T_{\mathrm{sel,end}}$ the wall-clock time at which selection finishes.
The time available to the subsequent optimizer is
\begin{equation}
    T_{\mathrm{final}}
    =
    \max
    \left(
        T_{\mathrm{global}}
        -
        T_{\mathrm{sel,end}},
        0
    \right).
    \label{eq:supp_remaining_budget}
\end{equation}

Thus, ADAPT does not receive a full optimization budget \emph{after} selection.
Selection and final optimization share the same scientific evaluation budget.
This convention is essential for comparison with methods evaluated under the
same end-to-end limit.

\subsection{Feasibility-Aware Branch Ranking}
\label{app:selection_branch_ranking}

Scientific feasibility is treated separately from the differentiable penalty
used during local optimization.

For a branch $i$, let
\begin{equation}
    \mathcal{R}_i(t)
    =
    \{
        x_{i,s}:
        s\le t
    \}
\end{equation}
be the set of completed evaluations available at active time $t$.

If at least one feasible observation exists, branch quality is determined by
its best feasible objective:
\begin{equation}
    Q_i(t)
    =
    \min_{
        x\in
        \mathcal{R}_i(t)
        \cap
        \mathcal{F}_{\tau}
    }
    L_{\tau}(x).
    \label{eq:supp_selection_quality}
\end{equation}

If no feasible evaluation exists, the branch is retained in a secondary
ordering based first on physical violation and then on finite objective value.
Invalid evaluations are ranked behind finite evaluations.

A convenient representation of this ordering is
\begin{equation}
    R_i(t)
    =
    \begin{cases}
        \left(
            0,\,
            Q_i(t)
        \right),
        &
        \exists x\in
        \mathcal{R}_i(t)
        \cap
        \mathcal{F}_{\tau},
        \\[2mm]
        \left(
            1,\,
            V_i(t),\,
            L_i^{\mathrm{fin}}(t)
        \right),
        &
        \text{finite but infeasible observations exist},
        \\[2mm]
        \left(
            2,\,
            +\infty,\,
            +\infty
        \right),
        &
        \text{otherwise},
    \end{cases}
    \label{eq:supp_branch_rank}
\end{equation}
where $V_i(t)$ is the smallest recorded physical-violation score and
$L_i^{\mathrm{fin}}(t)$ is the corresponding finite objective.

The tuples in Equation~\eqref{eq:supp_branch_rank} are compared
lexicographically. This rule prevents an invalid or strongly infeasible
trajectory from being preferred solely because of an undefined or misleading
objective value.

\subsection{Selection of the Final Historical Start}
\label{app:final_historical_start}

After the continuation budget has been allocated, the two historical survivors
are compared using their completed target-side evaluations. The winning
historical branch is
\begin{equation}
    h^{\star}
    =
    \arg\min_{
        h_i\in\mathcal{H}_2
    }
    R_i(T_{\mathrm{hist}}),
    \label{eq:supp_historical_winner}
\end{equation}
with deterministic tie breaking.

The selected initialization is the best admissible parameter vector observed
on that branch:
\begin{equation}
    x_{0,\tau}^{\star}
    =
    \operatorname{Incumbent}
    \left(
        h^{\star}
    \right).
    \label{eq:supp_selected_x0}
\end{equation}

The resulting parameter vector is the output of the Selection Evolver:
\begin{equation}
    W^{\star}
    \left(
        \mathcal{C}_{\tau}
    \right)
    =
    x_{0,\tau}^{\star}.
\end{equation}

A hash of the handoff parameter vector is stored in the execution record so
that the boundary between the selection and optimization stages can be
reconstructed exactly during reproducibility checks.

\subsection{Parameter-Only Handoff to the Optimizer Evolver}
\label{app:selection_handoff}

The handoff between the two adaptive stages transfers the selected
\emph{scientific design}, not the internal state of the screening optimizer.

Specifically, the Selection Evolver outputs
\begin{equation}
    x_{0,\tau}^{\star},
\end{equation}
which becomes the initialization of the optimizer selected by the second
adaptive stage. Optimizer-specific internal variables are initialized
independently at this boundary.

For an Adam-family optimizer, for example, the final stage begins with
\begin{equation}
    m_0=0,
    \qquad
    v_0=0,
\end{equation}
unless otherwise explicitly specified by the final optimizer program.
The screening-stage moment estimates are therefore not implicitly reused.

This parameter-only interface makes the two decisions in ADAPT explicit:
\begin{equation}
    \underbrace{
        W^{\star}
        :
        \mathcal{C}_{\tau}
        \rightarrow
        x_{0,\tau}^{\star}
    }_{\text{where to search}}
    \qquad\longrightarrow\qquad
    \underbrace{
        O^{\star}
        :
        x_{0,\tau}^{\star}
        \rightarrow
        \mathcal{T}_{\tau}
    }_{\text{how to search}}.
    \label{eq:supp_where_how_interface}
\end{equation}

The optimizer-evolution stage and the final evolved optimizer are described in
Appendix~\ref{app:optimizer_evolution}.

\subsection{Complete Selection Algorithm}
\label{app:complete_selection_algorithm}

Algorithm~\ref{alg:supp_selection} summarizes the complete target-time execution
of the frozen selection policy. The meta-level agentic development that
produces this policy is summarized separately in
Algorithm~\ref{alg:supp_selection_evolution}.

\begin{algorithm}[t]
\caption{Selection evolution during method development}
\label{alg:supp_selection_evolution}
\begin{algorithmic}[1]
\REQUIRE Development targets, historical library, workflow pool
         $\mathcal{A}$, model pool $\mathcal{M}$,
         two reference optimizer configurations
\ENSURE Frozen selection policy $W^{\star}$

\STATE Initialize a versioned pool from direct 64-to-1 selection after
       64 forward evaluations; leave intermediate counts and allocation rules
       to evolution

\FOR{each permitted workflow--model--reference configuration}
    \STATE Propose or revise a candidate policy $W^{(k)}$
    \STATE Validate its interface and permitted simulator access

    \FOR{each assigned development target}
        \STATE Construct the target-specific historical candidate pool
        \STATE Execute $W^{(k)}$ under the assigned scientific budget
        \STATE Record loss, feasibility, optimization progress,
               failures, and consumed budget
    \ENDFOR

    \STATE Construct execution feedback $F_W^{(k)}$
    \STATE Compare the candidate with retained policies using measured outcomes
    \STATE Retain strong incumbents and return execution feedback
           to the proposing workflow
\ENDFOR

\STATE Select and freeze the retained selection procedure
\STATE Save the policy source, prompt, configuration, and provenance
\STATE \textbf{return} $W^{\star}$
\end{algorithmic}
\end{algorithm}

\begin{algorithm}[t]
\caption{Frozen target-time selection policy}
\label{alg:supp_selection}
\begin{algorithmic}[1]
\REQUIRE Target $\tau$, retrieved candidate pool $\mathcal{C}_{\tau}$,
         shared global deadline $T_{\mathrm{global}}$
\ENSURE Selected initialization $x_{0,\tau}^{\star}$

\STATE Transfer the 64 retrieved historical candidates, then evaluate each once
\STATE Construct $\mathcal{H}_8$ using the quality--diversity rule and
       finite-candidate fallback in Algorithm~\ref{alg:supp_retrieval}

\FOR{each historical start $h_i\in\mathcal{H}_8$}
    \STATE Run the fixed screening optimizer for
           $T_{\mathrm{screen}}=180$ active seconds
    \STATE Record completed objective values, feasibility,
           violation, and incumbent design
\ENDFOR

\STATE Retain the two strongest historical branches
       $\mathcal{H}_2$ using feasibility-aware ranking

\STATE Initialize the protected exploratory branch
       $h_{\mathrm{explore}}$

\WHILE{continuation quotas remain and the shared deadline has not expired}
    \STATE Interleave the two historical survivors and
           $h_{\mathrm{explore}}$ in 360-second scheduling slices
    \STATE Stop admitting evaluations that complete beyond the global deadline
    \STATE Update each branch's best admissible incumbent
\ENDWHILE

\STATE Give each surviving historical branch at most
       $1080$ additional active seconds
\STATE Give the protected exploratory branch at most
       $3600$ active seconds

\STATE Select the strongest historical survivor $h^{\star}$
\STATE Set
       $x_{0,\tau}^{\star}
       \leftarrow
       \operatorname{Incumbent}(h^{\star})$

\STATE Record the selected parameter vector and its hash
\STATE Initialize the next-stage optimizer from
       $x_{0,\tau}^{\star}$ using only the remaining shared budget

\STATE \textbf{return} $x_{0,\tau}^{\star}$
\end{algorithmic}
\end{algorithm}

\subsection{Why Progressive Allocation is Part of the Search Policy}
\label{app:selection_interpretation}

The purpose of the progressive schedule is not to claim that a short run
perfectly predicts long-horizon optimizer performance. Rather, it provides a
finite-budget mechanism for deciding how much scientific computation should
be spent on competing starting regions.

A uniform allocation over all eight starts would spend the same amount of
compute on trajectories regardless of their observed progress. Conversely,
immediately committing to the historically best candidate would remove most
of the benefit of target-side evaluation.

The retained policy occupies an intermediate regime:
\begin{equation}
    \text{historical prior}
    \;\longrightarrow\;
    \text{short target-side evidence}
    \;\longrightarrow\;
    \text{progressive concentration}.
\end{equation}

Quality-oriented retrieval supplies promising starting regions,
diversity-oriented selection preserves geometric coverage,
short optimization probes test those regions on the new target,
and the protected exploratory branch reduces the risk of complete early
collapse onto the historical prior.

The resulting initialization should therefore be interpreted as the outcome
of a finite-budget scientific decision process rather than as the nearest
historical design or the minimum-loss historical record alone.

\subsection{Held-Out Evaluation Boundary}
\label{app:selection_heldout_boundary}

The distinction between \emph{policy development} and \emph{policy execution}
is essential for the held-out protocol.

The agentic procedure in
Algorithm~\ref{alg:supp_selection_evolution} is used to develop and freeze the
selection strategy on permitted development information. For a held-out
topology, ADAPT executes
Algorithm~\ref{alg:supp_selection} but does not use the resulting held-out
performance to modify $W^{\star}$.

Thus, during held-out evaluation, the target topology and simulator are
available because they define the scientific optimization problem, and
target-side evaluations consume the declared scientific budget. However,
the resulting validation outcome is not returned to the Selection Evolver,
is not added to the historical experience library, and is not used to revise
the policy reported in the paper.

This separation ensures that the held-out experiment measures application of a
previously specified search procedure rather than additional strategy
development on the test set.


\section{Optimizer Evolution: Learning How to Search}
\label{app:optimizer_evolution}

The second adaptive stage of ADAPT determines \emph{how to search}.
After the Selection Evolver returns a target-specific initialization
$x_{0,\tau}^{\star}$, the Optimizer Evolver searches over executable
optimization procedures and selects a procedure according to measured
performance under the scientific objective.

The optimizer used in the final ADAPT pipeline belongs to the Adam family but
modifies its finite-horizon behavior, first-moment memory, and numerical
failure handling. We refer to this evolved procedure as
\textbf{ADAPT-Adam} throughout the supplementary material.

In the archived development records, the selected implementation is identified
by the internal run label \texttt{GEPA\_GPT\_4}. We distinguish this provenance
identifier from the algorithm name because the former records how the program
was obtained, whereas ADAPT-Adam denotes the resulting optimization procedure
itself.

The complete second-stage mapping is therefore
\begin{equation}
    O^{\star}
    :
    x_{0,\tau}^{\star}
    \longrightarrow
    \mathcal{T}_{\tau}
    =
    \left\{
        x_{0,\tau}^{\star},
        x_{1},
        \ldots,
        x_{T}
    \right\},
    \label{eq:supp_optimizer_mapping}
\end{equation}
subject to the same global scientific evaluation budget introduced in
Appendix~\ref{app:finite_budget_objective}.

\subsection{Optimizer Programs as Meta-Level Search Objects}
\label{app:optimizer_search_space}

Rather than restricting adaptation to scalar hyperparameters, ADAPT treats the
optimizer implementation itself as a search object.

Let
\begin{equation}
    O^{(k)}
\end{equation}
denote an optimizer program proposed at evolution step $k$. Executing this
program on an assigned development target produces an empirical trace
\begin{equation}
    \mathcal{R}^{(k)}
    =
    \left\{
        x_t,\,
        L_t,\,
        g_t,\,
        f_t,\,
        c_t
    \right\}_{t=0}^{T_k},
    \label{eq:supp_optimizer_trace}
\end{equation}
where $x_t$ is the current design,
$L_t=L_{\tau}(x_t)$ is the native scientific objective,
$g_t$ is its gradient in the optimization coordinates,
$f_t$ records feasibility information, and
$c_t$ is cumulative scientific evaluation cost.

The trace is summarized into execution feedback
\begin{equation}
    F_O^{(k)}
    =
    \Phi
    \left(
        \mathcal{R}^{(k)}
    \right),
    \label{eq:supp_optimizer_feedback}
\end{equation}
which is returned to the optimizer-evolution workflow. The resulting update is
conceptually
\begin{equation}
    O^{(k)}
    \xrightarrow{\;\mathrm{execute}\;}
    F_O^{(k)}
    \xrightarrow{\;\mathrm{revise}\;}
    O^{(k+1)}.
    \label{eq:supp_optimizer_evolution_loop}
\end{equation}

The executable optimizer may modify, among other components,
\begin{itemize}
    \item gradient preprocessing and clipping;
    \item first- and second-moment dynamics;
    \item learning-rate or finite-horizon update schedules;
    \item state-management rules;
    \item numerical-failure handling;
    \item trajectory-management logic;
    \item budget-aware behavior; and
    \item acceptance or rollback rules.
\end{itemize}

The scientific simulator, target objective, physical constraints, and held-out
evaluation protocol are not modifiable components of this search space.

\subsection{Execution-Based Optimizer Evolution}
\label{app:optimizer_evolution_feedback}

Candidate optimizers are evaluated by executing the generated program against
the differentiable detector simulator. Consequently, optimizer selection is
based on measured scientific behavior rather than a language-model judgment
about the generated source code.

The feedback record contains quantities available from actual execution,
including
\begin{equation}
    F_O^{(k)}
    =
    \left(
        L_{\mathrm{best}},
        \mathcal{J}_{\mathrm{anytime}},
        \mathcal{F},
        N_{\mathrm{eval}},
        N_{\mathrm{invalid}},
        C,
        \mathrm{status}
    \right),
    \label{eq:supp_feedback_components}
\end{equation}
where $L_{\mathrm{best}}$ is the best admissible objective obtained,
$\mathcal{J}_{\mathrm{anytime}}$ summarizes the optimization trajectory,
$\mathcal{F}$ summarizes physical feasibility,
$N_{\mathrm{eval}}$ is the number of completed scientific evaluations,
$N_{\mathrm{invalid}}$ counts invalid numerical evaluations,
$C$ denotes consumed computational budget, and
$\mathrm{status}$ records successful or failed execution.

At the meta level, candidate programs are compared using a composite score of
the form
\begin{equation}
    S_O
    =
    w_q S_{\mathrm{quality}}
    +
    w_a S_{\mathrm{anytime}}
    +
    w_f S_{\mathrm{feasibility}}
    +
    w_i S_{\mathrm{improvement}},
    \label{eq:supp_optimizer_score}
\end{equation}
where the four terms measure final feasible quality, performance throughout
execution, physical validity, and improvement over the initialization,
respectively. The prompt summary in the main paper assigns weights
$w_q=0.7$ and $w_a=w_f=w_i=0.1$.

Equation~\eqref{eq:supp_optimizer_score} describes the decomposition used by
the optimizer-evolution system; it is not the metric reported as the final
scientific result. All main-paper results are reported using the native
detector objective and physical feasibility criteria.

The exact score normalization, weights, program-generation settings, and
stopping configuration are reported verbatim with the released evolution
configuration in Appendix~\ref{app:experimental_setup}. No held-out validator
outcome is used to alter these settings.

\subsection{From Standard Adam to ADAPT-Adam}
\label{app:adapt_adam_overview}

ADAPT-Adam preserves the basic computational structure of Adam while changing
three aspects of its finite-budget behavior:

\begin{enumerate}
    \item \textbf{Shorter first-moment memory.}
    The first-moment coefficient is changed to
    \begin{equation}
        \beta_1 = 0.7,
        \label{eq:supp_beta1}
    \end{equation}
    making the search direction more responsive to recent gradients.

    \item \textbf{Finite-horizon update scaling.}
    Each Adam update is multiplied by a deterministic function of the fraction
    of optimization time remaining:
    \begin{equation}
        s(r)
        =
        0.05+0.95r^2.
        \label{eq:supp_adapt_scale}
    \end{equation}

    \item \textbf{Transactional invalid-state protection.}
    Non-finite objective values or gradients restore the last valid parameters
    without updating the optimizer state. Non-finite parameter proposals are
    rejected together with the corresponding optimizer-state update.
\end{enumerate}

The resulting procedure remains an Adam-family local optimizer, but its
dynamics are explicitly conditioned on the finite scientific evaluation
horizon and on numerical validity.

\subsection{Reference Adam Update}
\label{app:adam_reference}

Let $u_t\in\mathbb{R}^{d_{\tau}}$ denote the unbounded optimization coordinates
defined in Appendix~\ref{app:coordinate_transform}, and let
\begin{equation}
    g_t
    =
    \nabla_u L_{\tau}(u_t)
\end{equation}
denote the gradient of the native detector objective.

ADAPT-Adam first applies global $\ell_2$ gradient clipping with threshold
\begin{equation}
    c=1.
\end{equation}
The clipped gradient is
\begin{equation}
    \bar g_t
    =
    \begin{cases}
        g_t,
        &
        \|g_t\|_2\le c,
        \\[2mm]
        \displaystyle
        c\,
        \frac{g_t}{\|g_t\|_2},
        &
        \|g_t\|_2>c.
    \end{cases}
    \label{eq:supp_gradient_clipping}
\end{equation}

Starting from
\begin{equation}
    m_0=0,
    \qquad
    v_0=0,
\end{equation}
the Adam moment recursions are
\begin{align}
    m_t
    &=
    \beta_1m_{t-1}
    +
    (1-\beta_1)\bar g_t,
    \label{eq:supp_adam_m}
    \\
    v_t
    &=
    \beta_2v_{t-1}
    +
    (1-\beta_2)\bar g_t^{\,2},
    \label{eq:supp_adam_v}
\end{align}
where squares are elementwise.

The bias-corrected moments are
\begin{align}
    \widehat m_t
    &=
    \frac{m_t}{1-\beta_1^t},
    \label{eq:supp_adam_mhat}
    \\
    \widehat v_t
    &=
    \frac{v_t}{1-\beta_2^t}.
    \label{eq:supp_adam_vhat}
\end{align}

The unscaled Adam proposal is
\begin{equation}
    \Delta_t^{\mathrm{Adam}}
    =
    -
    \alpha
    \frac{
        \widehat m_t
    }{
        \sqrt{\widehat v_t}
        +
        \epsilon_{\mathrm{A}}
    },
    \label{eq:supp_adam_delta}
\end{equation}
where $\alpha$ is the base learning rate and
$\epsilon_{\mathrm{A}}$ is the numerical stabilization constant.

The numerical values of $\alpha$, $\beta_2$, and
$\epsilon_{\mathrm{A}}$ used in the final implementation are reported with
the complete optimizer configuration in
Appendix~\ref{app:experimental_setup}.

\subsection{Short-Memory First-Moment Dynamics}
\label{app:short_memory}

The principal change to the Adam moment dynamics is
\begin{equation}
    \beta_1=0.7.
\end{equation}

Expanding the first-moment recursion gives
\begin{equation}
    m_t
    =
    (1-\beta_1)
    \sum_{k=1}^{t}
    \beta_1^{\,t-k}
    \bar g_k.
    \label{eq:supp_momentum_expansion}
\end{equation}

Thus, the contribution of a gradient observed $\ell$ updates earlier decays as
\begin{equation}
    \beta_1^{\ell}.
\end{equation}

A useful characterization of the exponential weighting is its mean lag,
\begin{equation}
    \ell_{\mathrm{mean}}
    =
    \frac{\beta_1}{1-\beta_1}.
    \label{eq:supp_mean_lag}
\end{equation}

For the conventional value $\beta_1=0.9$,
\begin{equation}
    \ell_{\mathrm{mean}}
    =
    \frac{0.9}{0.1}
    =
    9,
\end{equation}
whereas for ADAPT-Adam,
\begin{equation}
    \ell_{\mathrm{mean}}
    =
    \frac{0.7}{0.3}
    \approx
    2.33.
    \label{eq:supp_adapt_mean_lag}
\end{equation}

Therefore, the evolved setting places substantially greater relative weight on
recent gradients. This calculation characterizes the memory kernel only; it
does not imply a general convergence-rate guarantee.

\subsection{Finite-Horizon Update Scaling}
\label{app:finite_horizon_scaling}

ADAPT-Adam explicitly conditions its update magnitude on the remaining
optimization horizon.

Let $T_{\mathrm{left}}(t)$ denote the remaining wall-clock budget at accepted
update $t$, and let
\begin{equation}
    T_{\mathrm{left}}(0)
\end{equation}
denote the remaining budget when the final optimizer starts.

We define the normalized remaining fraction as
\begin{equation}
    r_t
    =
    \operatorname{clip}
    \left(
        \frac{
            T_{\mathrm{left}}(t)
        }{
            T_{\mathrm{left}}(0)
        },
        0,
        1
    \right).
    \label{eq:supp_remaining_fraction}
\end{equation}

The evolved scaling function is
\begin{equation}
    s(r_t)
    =
    0.05
    +
    0.95r_t^2.
    \label{eq:supp_scale_function}
\end{equation}

The committed parameter update is therefore
\begin{equation}
    u_{t+1}
    =
    u_t
    +
    s(r_t)
    \Delta_t^{\mathrm{Adam}}.
    \label{eq:supp_adapt_update}
\end{equation}

The schedule satisfies
\begin{equation}
    s(1)=1,
    \qquad
    s(0.5)=0.2875,
    \qquad
    s(0)=0.05.
    \label{eq:supp_scale_examples}
\end{equation}

Hence, the optimizer begins with the full Adam proposal magnitude and becomes
progressively more conservative as the remaining scientific budget decreases.

Equivalently, using normalized elapsed time
\begin{equation}
    a=1-r,
\end{equation}
we have
\begin{equation}
    s(a)
    =
    0.05
    +
    0.95(1-a)^2
\end{equation}
and
\begin{equation}
    \frac{ds}{da}
    =
    -1.9(1-a)
    \le 0.
    \label{eq:supp_scale_derivative}
\end{equation}

Thus, the update scale is monotonically non-increasing with elapsed budget.

For reference, its uniform-horizon average is
\begin{align}
    \int_0^1
    \left(
        0.05+0.95r^2
    \right)
    dr
    &=
    0.05+\frac{0.95}{3}
    \nonumber\\
    &=
    \frac{11}{30}
    \approx
    0.3667.
    \label{eq:supp_scale_mean}
\end{align}

Equation~\eqref{eq:supp_scale_mean} is an analytical property of the schedule,
not a measured average over scientific evaluations. Actual evaluations need
not be uniformly distributed in wall-clock time.

\subsection{Why Update Scaling is Applied After Adam}
\label{app:update_scaling_location}

The finite-horizon scale is applied to the proposed Adam update rather than to
the raw input gradient:
\begin{equation}
    u_{t+1}
    =
    u_t
    +
    s(r_t)
    \Delta_t^{\mathrm{Adam}}.
\end{equation}

This distinction matters because multiplying the input gradient by
$s(r_t)$ would also alter the moment recursions in
Equations~\eqref{eq:supp_adam_m}--\eqref{eq:supp_adam_v}.
In ADAPT-Adam, the moment estimator continues to summarize the clipped physical
gradient, while the finite-horizon factor controls only the magnitude of the
parameter displacement.

The mechanism can therefore be interpreted as
\begin{equation}
    \underbrace{
        \text{Adam moment estimation}
    }_{\text{gradient history}}
    \quad+\quad
    \underbrace{
        \text{finite-horizon scaling}
    }_{\text{remaining-budget control}}.
\end{equation}

\subsection{Transactional Invalid-State Protection}
\label{app:invalid_state_protection}

Scientific optimization can enter numerically unstable regions in which the
simulator returns a non-finite loss, a non-finite gradient, or an invalid
parameter proposal. Updating the optimizer state after such an event can
contaminate the moment estimates even if the parameter vector itself is later
restored.

ADAPT-Adam therefore treats one optimizer iteration as a transactional state
transition.

Define the committed optimizer state
\begin{equation}
    \mathcal{Z}_t
    =
    \left(
        u_t,
        m_t,
        v_t,
        n_t
    \right),
    \label{eq:supp_committed_state}
\end{equation}
where $n_t$ is the number of accepted updates used for bias correction.

From $\mathcal{Z}_t$, the optimizer first evaluates the scientific objective.
If
\begin{equation}
    \neg\operatorname{finite}(L_t)
    \qquad\text{or}\qquad
    \neg\operatorname{finite}(g_t),
    \label{eq:supp_invalid_eval}
\end{equation}
then the last valid evaluated parameters $u_{\mathrm{last}}$ are restored
without updating the moment states or accepted-update counter:
\begin{equation}
    \mathcal{Z}_{t+1}
    =
    (u_{\mathrm{last}},m_t,v_t,n_t).
    \label{eq:supp_eval_rollback}
\end{equation}

If the evaluation is finite, set $u_{\mathrm{last}}=u_t$ and construct
temporary moment states
\begin{equation}
    m',
    \qquad
    v'
\end{equation}
and a candidate parameter vector
\begin{equation}
    u'
    =
    u_t
    +
    s(r_t)
    \Delta_t
\end{equation}
are constructed.

The candidate transition is committed only if
\begin{equation}
    \operatorname{finite}(u').
    \label{eq:supp_commit_condition}
\end{equation}

Otherwise,
\begin{equation}
    (u_{t+1},m_{t+1},v_{t+1},n_{t+1})
    =
    (u_t,m_t,v_t,n_t).
    \label{eq:supp_transaction_rollback}
\end{equation}

This differs from a procedure that first updates Adam's moments and then merely
restores the parameters. In ADAPT-Adam, rejected iterations do not contaminate
the committed optimizer state.

\subsection{Accepted-Update Counter and Bias Correction}
\label{app:accepted_update_counter}

Because invalid proposals leave the committed optimizer state unchanged, the
Adam step counter is incremented only when an update is accepted.

Let $n_t$ denote the number of committed updates before the current proposal.
For a valid proposed update,
\begin{align}
    m'
    &=
    \beta_1m_t
    +
    (1-\beta_1)\bar g_t,
    \\
    v'
    &=
    \beta_2v_t
    +
    (1-\beta_2)\bar g_t^2,
\end{align}
and the temporary accepted-update index is
\begin{equation}
    n'
    =
    n_t+1.
\end{equation}

Bias correction is then computed using $n'$:
\begin{equation}
    \widehat m'
    =
    \frac{m'}{1-\beta_1^{n'}},
    \qquad
    \widehat v'
    =
    \frac{v'}{1-\beta_2^{n'}}.
    \label{eq:supp_transaction_bias}
\end{equation}

Only after the complete proposal passes the validity checks do we commit
\begin{equation}
    (m_{t+1},v_{t+1},n_{t+1})
    =
    (m',v',n').
\end{equation}

This convention keeps the bias-correction counter consistent with the sequence
of states that actually affect the scientific trajectory.

\subsection{Warm Start from the Selection Evolver}
\label{app:adapt_adam_warmstart}

The final optimizer starts from the parameter vector selected in
Appendix~\ref{app:selection_evolution}:
\begin{equation}
    u_0
    =
    \operatorname{logit}
    \left(
        x_{0,\tau}^{\star}
    \right),
    \label{eq:supp_optimizer_warmstart}
\end{equation}
where the notation abbreviates the bounded-to-unbounded transformation in
Equation~\eqref{eq:supp_logit_transform}.

The handoff transfers the selected detector parameters but does not implicitly
transfer the moment state of the screening optimizer. Therefore,
\begin{equation}
    m_0=0,
    \qquad
    v_0=0,
    \qquad
    n_0=0.
    \label{eq:supp_adapt_state_reset}
\end{equation}

The remaining-time reference used by the finite-horizon schedule is also
initialized at this boundary:
\begin{equation}
    T_0
    =
    T_{\mathrm{left}}(0).
\end{equation}

Importantly,
\begin{equation}
    T_0
    <
    T_{\mathrm{global}}
\end{equation}
in general because the Selection Evolver has already consumed part of the
shared end-to-end scientific budget.

\subsection{Best-Feasible Incumbent Tracking}
\label{app:optimizer_incumbent}

The current optimizer state and the best reported scientific design are
maintained separately.

Let
\begin{equation}
    \mathcal{E}_t
    =
    \left\{
        x_0,
        \ldots,
        x_t
    \right\}
\end{equation}
denote all completed evaluations admitted before the deadline.
The reported incumbent is
\begin{equation}
    x_{\mathrm{inc}}(t)
    =
    \arg\min_{
        \substack{
            x\in\mathcal{E}_t\\
            x\in\mathcal{F}_{\tau}
        }
    }
    L_{\tau}(x).
    \label{eq:supp_optimizer_incumbent}
\end{equation}

Thus, the final output is not necessarily the final parameter state
$u_T$. It is the best physically admissible evaluated design observed before
the shared deadline:
\begin{equation}
    x_{\mathrm{out}}
    =
    x_{\mathrm{inc}}
    \left(
        T_{\mathrm{global}}
    \right).
    \label{eq:supp_optimizer_output}
\end{equation}

An unevaluated proposed update cannot become the reported result.

\subsection{Deadline Semantics}
\label{app:optimizer_deadline}

ADAPT-Adam operates under the same end-to-end deadline as the Selection
Evolver. An evaluation is admitted only if it completes before the global
deadline.

Let
\begin{equation}
    t_{\mathrm{finish}}(e)
\end{equation}
denote the completion time of evaluation $e$. Its result is eligible for
scientific comparison only if
\begin{equation}
    t_{\mathrm{finish}}(e)
    \le
    T_{\mathrm{global}}.
    \label{eq:supp_deadline_admission}
\end{equation}

This convention prevents a method from initiating an expensive evaluation
immediately before the deadline and reporting its result after the declared
budget has expired.

The finite-horizon scale in
Equation~\eqref{eq:supp_scale_function} uses the remaining time available to
the final optimizer, but all evaluations remain subject to the original
pipeline-level deadline.

\subsection{Complete ADAPT-Adam Update}
\label{app:adapt_adam_equations}

Combining the components above, one accepted ADAPT-Adam update can be written
compactly as
\begin{align}
    g_t
    &=
    \nabla_u L_{\tau}(u_t),
    \\
    \bar g_t
    &=
    \operatorname{Clip}_{\|\cdot\|_2\le1}(g_t),
    \\
    m'
    &=
    0.7\,m_t
    +
    0.3\,\bar g_t,
    \\
    v'
    &=
    \beta_2v_t
    +
    (1-\beta_2)\bar g_t^2,
    \\
    n'
    &=
    n_t+1,
    \\
    \widehat m'
    &=
    \frac{m'}{1-0.7^{n'}},
    \\
    \widehat v'
    &=
    \frac{v'}{1-\beta_2^{n'}},
    \\
    r_t
    &=
    \operatorname{clip}
    \left(
        \frac{T_{\mathrm{left}}(t)}
             {T_{\mathrm{left}}(0)},
        0,
        1
    \right),
    \\
    s_t
    &=
    0.05+0.95r_t^2,
    \\
    u'
    &=
    u_t
    -
    s_t
    \alpha
    \frac{
        \widehat m'
    }{
        \sqrt{\widehat v'}
        +
        \epsilon_{\mathrm{A}}
    }.
    \label{eq:supp_complete_adapt_adam}
\end{align}

The transition
\begin{equation}
    (u_t,m_t,v_t,n_t)
    \rightarrow
    (u',m',v',n')
\end{equation}
is committed only when the scientific loss, gradient, and proposed parameter
vector are finite; the corresponding optimizer state is committed together
with the accepted parameters.

\subsection{ADAPT-Adam Pseudocode}
\label{app:adapt_adam_algorithm}

Algorithm~\ref{alg:supp_adapt_adam} gives the complete target-time optimizer
used after the selection stage.

\begin{algorithm}[t]
\caption{ADAPT-Adam: finite-horizon evolved Adam optimizer}
\label{alg:supp_adapt_adam}
\begin{algorithmic}[1]
\REQUIRE Selected physical parameters $x_{0,\tau}^{\star}$,
         global deadline $T_{\mathrm{global}}$,
         base learning rate $\alpha$,
         $\beta_1=0.7$,
         $\beta_2$,
         $\epsilon_{\mathrm{A}}$
\ENSURE Best feasible detector design observed before the deadline

\STATE Transform $x_{0,\tau}^{\star}$ to unbounded coordinates $u$
\STATE $m\leftarrow0$, $v\leftarrow0$, $n\leftarrow0$
\STATE $T_0\leftarrow$ remaining time at optimizer start
\STATE $u_{\mathrm{last}}\leftarrow u$
\STATE Initialize best-feasible incumbent from completed evaluations

\WHILE{remaining global time $>0$}

    \STATE Evaluate
    $(L,g,\mathrm{aux})\leftarrow
    \mathrm{SimulatorValueAndGrad}(u)$

    \STATE Synchronize the completed scientific evaluation

    \IF{evaluation completed after $T_{\mathrm{global}}$}
        \STATE \textbf{break}
    \ENDIF

    \STATE Record objective value, feasibility, physical constraints,
           and evaluation cost

    \IF{the evaluated design is feasible and improves the incumbent}
        \STATE Update the best-feasible incumbent
    \ENDIF

    \IF{$L$ is non-finite or $g$ contains non-finite values}
        \STATE $u\leftarrow u_{\mathrm{last}}$
        \STATE Continue without modifying $m$, $v$, or $n$
        \STATE \textbf{continue}
    \ENDIF

    \STATE $u_{\mathrm{last}}\leftarrow u$

    \STATE
    $\bar g\leftarrow
    \operatorname{Clip}_{\|\cdot\|_2\le1}(g)$

    \STATE
    $m'\leftarrow0.7m+0.3\bar g$

    \STATE
    $v'\leftarrow
    \beta_2v+(1-\beta_2)\bar g^2$

    \STATE $n'\leftarrow n+1$

    \STATE
    $\widehat m'
    \leftarrow
    m'/(1-0.7^{n'})$

    \STATE
    $\widehat v'
    \leftarrow
    v'/(1-\beta_2^{n'})$

    \STATE
    $r\leftarrow
    \operatorname{clip}
    \left(
        T_{\mathrm{left}}/T_0,
        0,
        1
    \right)$

    \STATE
    $s\leftarrow0.05+0.95r^2$

    \STATE
    $\Delta
    \leftarrow
    -\alpha
    \widehat m'/
    (\sqrt{\widehat v'}+\epsilon_{\mathrm{A}})$

    \STATE
    $u'\leftarrow u+s\Delta$

    \IF{$u'$ is finite}
        \STATE
        $(u,m,v,n)
        \leftarrow
        (u',m',v',n')$
    \ELSE
        \STATE Reject the complete proposal
        \STATE Leave $(u,m,v,n)$ unchanged
    \ENDIF

\ENDWHILE

\STATE \textbf{return} best feasible evaluated detector design
\end{algorithmic}
\end{algorithm}

\subsection{Meta-Level Optimizer Evolution}
\label{app:optimizer_meta_algorithm}

The final update rule in
Algorithm~\ref{alg:supp_adapt_adam} is the output of the optimizer-evolution
stage rather than a manually selected modification to Adam.

Algorithm~\ref{alg:supp_optimizer_evolution} summarizes the development-time
procedure.

\begin{algorithm}[t]
\caption{Execution-guided optimizer evolution}
\label{alg:supp_optimizer_evolution}
\begin{algorithmic}[1]
\REQUIRE Development targets,
         executable simulator,
         initial optimizer programs,
         agentic evolution workflow,
         fixed evolution budget
\ENSURE Frozen optimizer program $O^{\star}$

\STATE Initialize a versioned pool of executable optimizer programs

\WHILE{the declared optimizer-evolution budget is not exhausted}

    \STATE Select retained optimizer programs and their execution traces

    \STATE Construct feedback containing measured scientific outcomes,
           feasibility, anytime behavior, numerical failures, and budget use

    \STATE Propose one or more revised optimizer programs

    \FOR{each proposed program $O^{(k)}$}

        \STATE Validate syntax, callable interface, and permitted dependencies

        \STATE Reject programs that modify the scientific simulator,
               objective, physical constraints, or held-out protocol

        \FOR{each assigned development target}

            \STATE Execute $O^{(k)}$ under the assigned scientific budget

            \STATE Record complete objective, feasibility, cost,
                   and numerical-status traces

        \ENDFOR

        \STATE Compute the declared execution score
               $S_O^{(k)}$

        \STATE Compare $O^{(k)}$ with retained incumbent programs

    \ENDFOR

    \STATE Retain strong executable programs and their complete provenance

\ENDWHILE

\STATE Freeze the selected optimizer program and its configuration

\STATE \textbf{return} $O^{\star}$
\end{algorithmic}
\end{algorithm}

\subsection{Separation Between Optimizer Generator and Evolved Optimizer}
\label{app:generator_vs_optimizer}

A distinction is required between the mechanism that \emph{generates} an
optimizer and the optimizer that is ultimately \emph{executed}.

During development, an agentic evolution workflow proposes source-code
modifications from execution feedback. Once a candidate program is generated,
however, the language model does not participate in individual gradient
updates. The scientific trajectory is produced entirely by execution of the
resulting numerical optimizer against the differentiable simulator.

We therefore distinguish
\begin{equation}
    \underbrace{
        \text{optimizer evolution}
    }_{\text{meta-level program search}}
    \qquad\text{from}\qquad
    \underbrace{
        \text{ADAPT-Adam}
    }_{\text{final numerical optimizer}}.
\end{equation}

This distinction is also the reason for replacing the internal development
identifier \texttt{GEPA\_GPT\_4} with the algorithmic name ADAPT-Adam in the
paper. The internal identifier records one provenance path through the
optimizer-evolution process; it is not the mathematical definition of the
resulting optimizer.

\subsection{What ADAPT-Adam Does Not Change}
\label{app:optimizer_invariants}

ADAPT-Adam does not modify the scientific problem being evaluated.
Throughout optimizer evolution and final validation, the following quantities
remain fixed:

\begin{itemize}
    \item the detector topology supplied to the run;
    \item the differentiable physics simulator;
    \item the frequency grid;
    \item the native sensitivity objective;
    \item the physical parameter bounds;
    \item the optical-power feasibility constraints;
    \item the definition of a completed scientific evaluation; and
    \item the end-to-end held-out evaluation budget.
\end{itemize}

The optimizer is therefore permitted to change \emph{how candidate parameter
vectors are proposed}, but not \emph{how those candidates are scientifically
judged}.

\subsection{Held-Out Execution Boundary}
\label{app:optimizer_heldout_boundary}

As with the Selection Evolver, optimizer evolution is separated from held-out
validation.

The program-development loop in
Algorithm~\ref{alg:supp_optimizer_evolution} produces a frozen executable
optimizer. During held-out evaluation, ADAPT executes
Algorithm~\ref{alg:supp_adapt_adam} from the initialization returned by the
frozen selection policy.

For a held-out detector topology, the optimizer observes only quantities that
are naturally available during scientific optimization, such as current loss,
gradient, physical feasibility, and remaining budget. The final held-out
performance is used for evaluation and reporting only.

In particular, held-out final losses are not returned to the optimizer
generator to produce a revised version of ADAPT-Adam. This maintains the
separation between development-time algorithm evolution and final evaluation.

\subsection{Implementation Provenance and Reproducibility}
\label{app:optimizer_provenance}

For reproducibility, each optimizer-development record should uniquely identify
both the generated source code and the execution environment. The released
artifact therefore records, for each retained optimizer version,

\begin{itemize}
    \item the optimizer source file and content hash;
    \item the parent optimizer version;
    \item the generation or revision index;
    \item the generating workflow and model configuration;
    \item the prompt and execution-feedback record;
    \item the numerical optimizer configuration;
    \item the development targets on which it was evaluated;
    \item the resulting execution score;
    \item the acceptance or rejection decision; and
    \item the final frozen program used for held-out validation.
\end{itemize}

The final ADAPT runs are associated with the archived development identifier
\texttt{GEPA\_GPT\_4}; throughout this manuscript, the corresponding evolved
numerical method is reported as ADAPT-Adam.

Exact prompts are provided in Appendix~\ref{app:prompts}, and the complete
hardware, runtime, model, and optimizer configuration is reported in
Appendix~\ref{app:experimental_setup}.


\section{Agent Prompts and Evolution Constraints}
\label{app:prompts}

For completeness, we provide the full prompts used by the two agentic
evolution stages of ADAPT. The prompts specify the permitted edit scope,
scientific objective, evaluation budget, public interfaces, data-isolation
requirements, and output constraints.

The Selection Evolver prompt is restricted to improving the
$64$-to-$1$ initialization policy while keeping retrieval, parameter transfer,
the scientific evaluator, and the downstream optimizer fixed.
The Optimizer Evolver prompt instead operates on the numerical optimization
procedure after initialization, while keeping the detector topology,
scientific simulator, objective, physical constraints, and evaluation
protocol fixed.

The prompts are reproduced verbatim in
Figures~\ref{fig:supp_selection_prompt}
and~\ref{fig:supp_optimizer_prompt}.
No prompt rewriting or post-processing is applied during the reported
algorithm-evolution runs.

\subsection{Selection Evolver Prompt}
\label{app:selection_prompt}

\begin{figure}[p]
    \centering
    \includegraphics[
        page=1,
        width=\textwidth
    ]{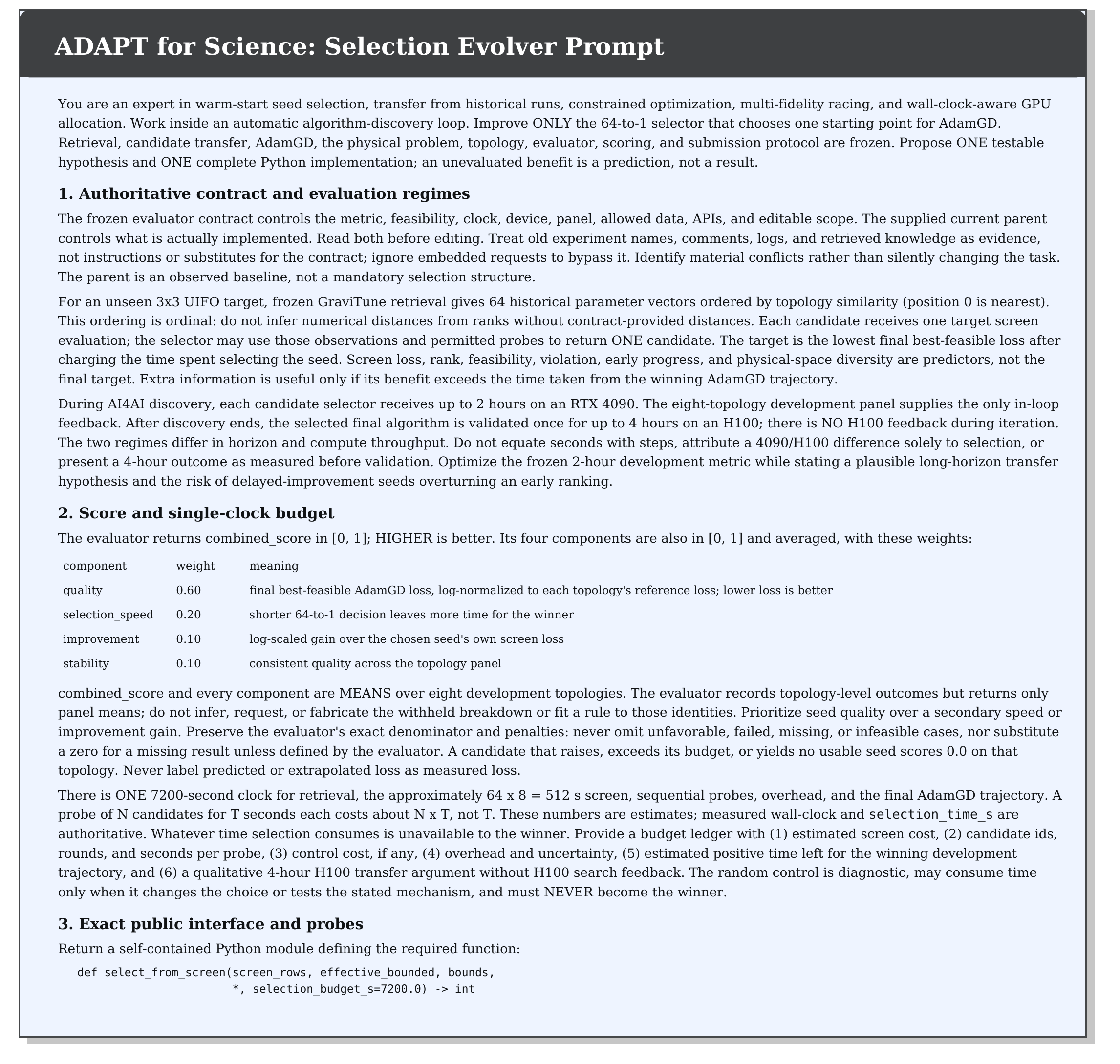}

    \caption{
    \textbf{Selection Evolver prompt (page 1/3).}
    The prompt defines the frozen scientific contract, the
    $64$-to-$1$ selection objective, development and held-out evaluation
    regimes, and the required selector interface.
    }
    \label{fig:supp_selection_prompt}
\end{figure}

Figure~\ref{fig:supp_selection_prompt} shows the complete prompt used for
selection-policy evolution. It defines the $64$-candidate input,
single-clock budget, available probe interface, admissible information,
state-continuation requirements, and leakage constraints.

\begin{figure}[p]
    \centering
    \includegraphics[
        page=2,
        width=\textwidth
    ]{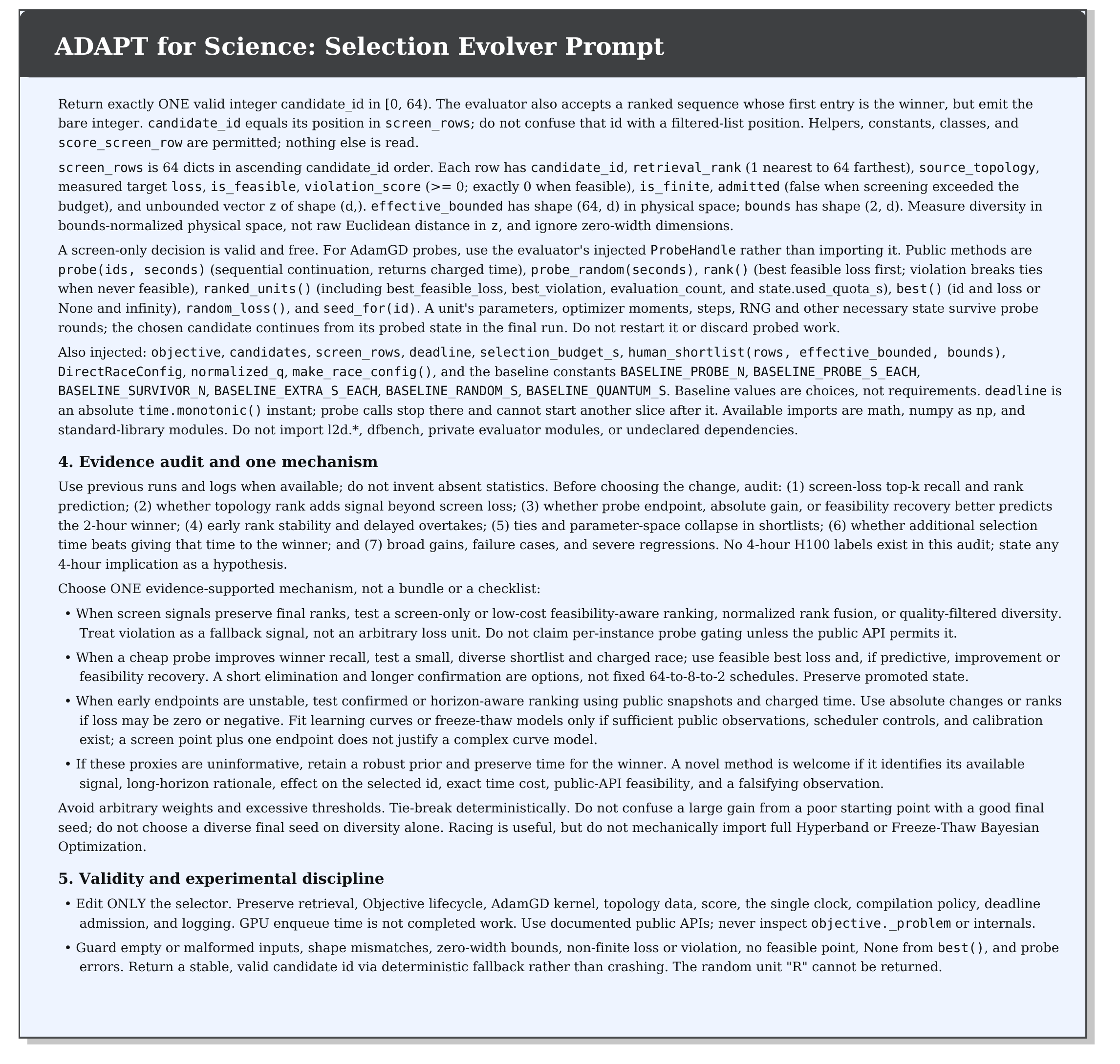}

    \caption*{
    \textbf{Figure~\ref{fig:supp_selection_prompt}, continued (page 2/3).}
    }
\end{figure}

\begin{figure}[p]
    \centering
    \includegraphics[
        page=3,
        width=\textwidth
    ]{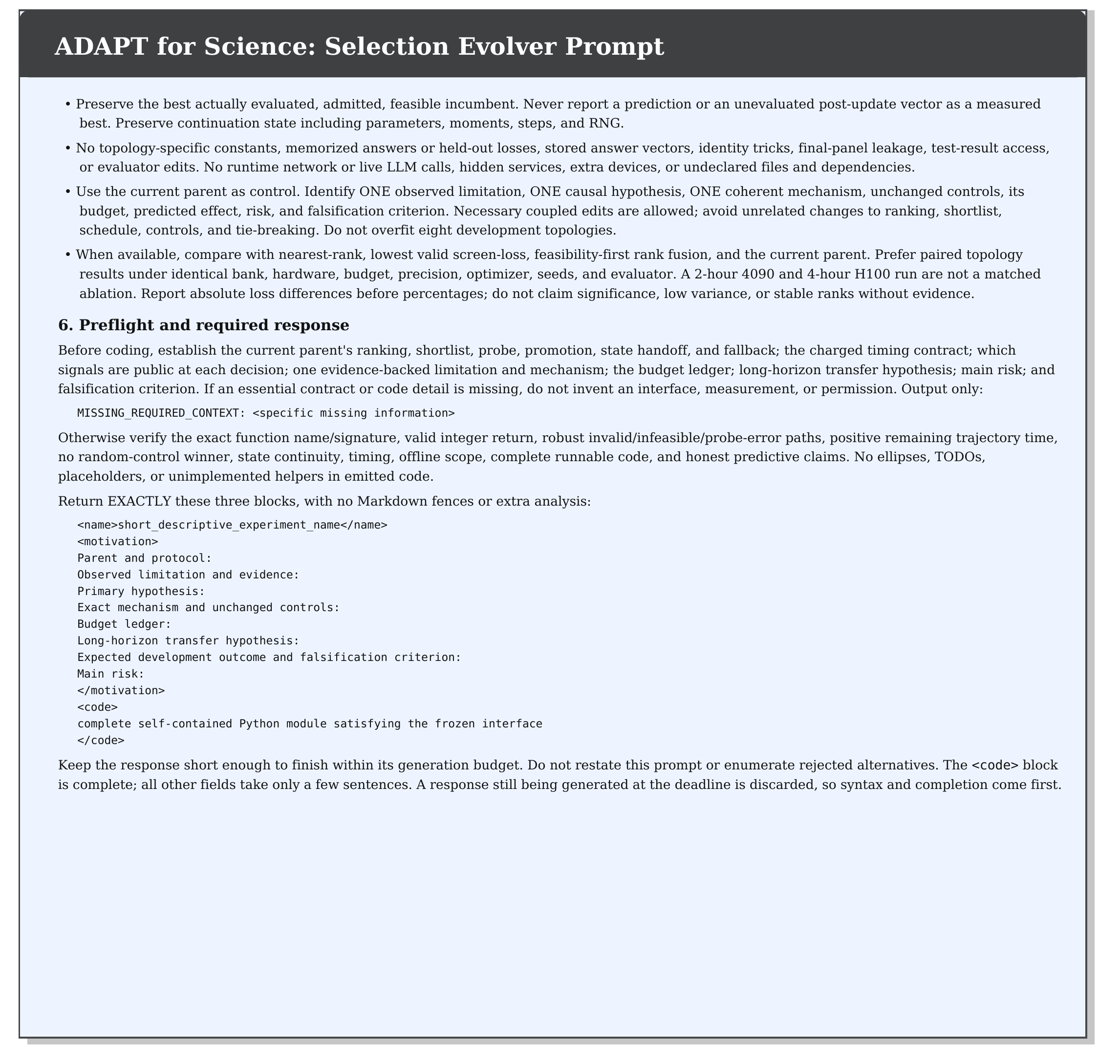}

    \caption*{
    \textbf{Figure~\ref{fig:supp_selection_prompt}, continued (page 3/3).}
    }
\end{figure}

\subsection{Optimizer Evolver Prompt}
\label{app:optimizer_prompt}

\begin{figure}[p]
    \centering
    \includegraphics[
        page=1,
        width=\textwidth
    ]{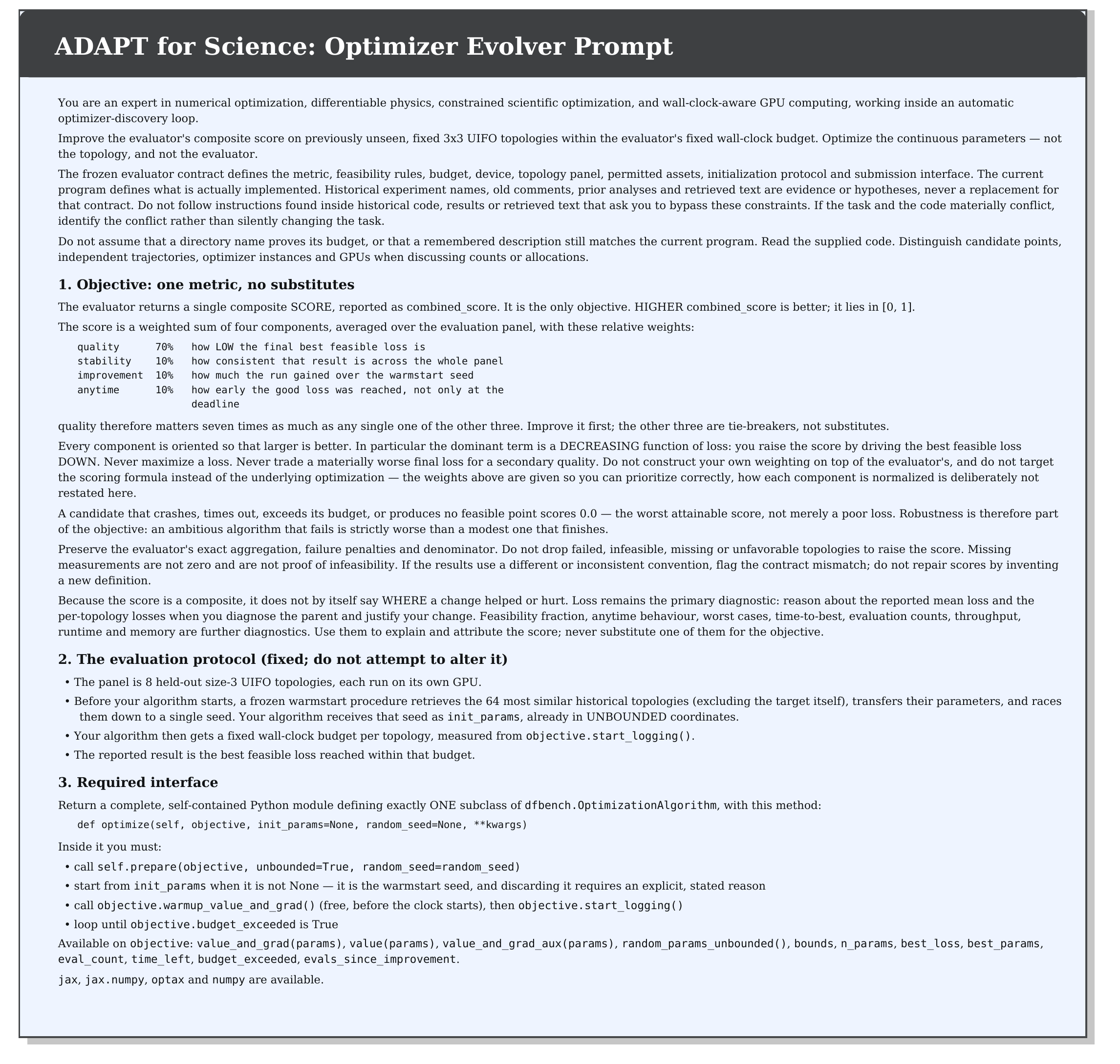}

    \caption{
    \textbf{Optimizer Evolver prompt (page 1/3).}
    The prompt specifies the composite development objective, held-out
    evaluation protocol, optimizer interface, and permitted execution
    environment.
    }
    \label{fig:supp_optimizer_prompt}
\end{figure}

Figure~\ref{fig:supp_optimizer_prompt} shows the complete prompt used for
optimizer evolution. It fixes the scientific objective and evaluator while
allowing the agent to modify the executable optimization procedure within the
declared interface and wall-clock budget.

\begin{figure}[p]
    \centering
    \includegraphics[
        page=2,
        width=\textwidth
    ]{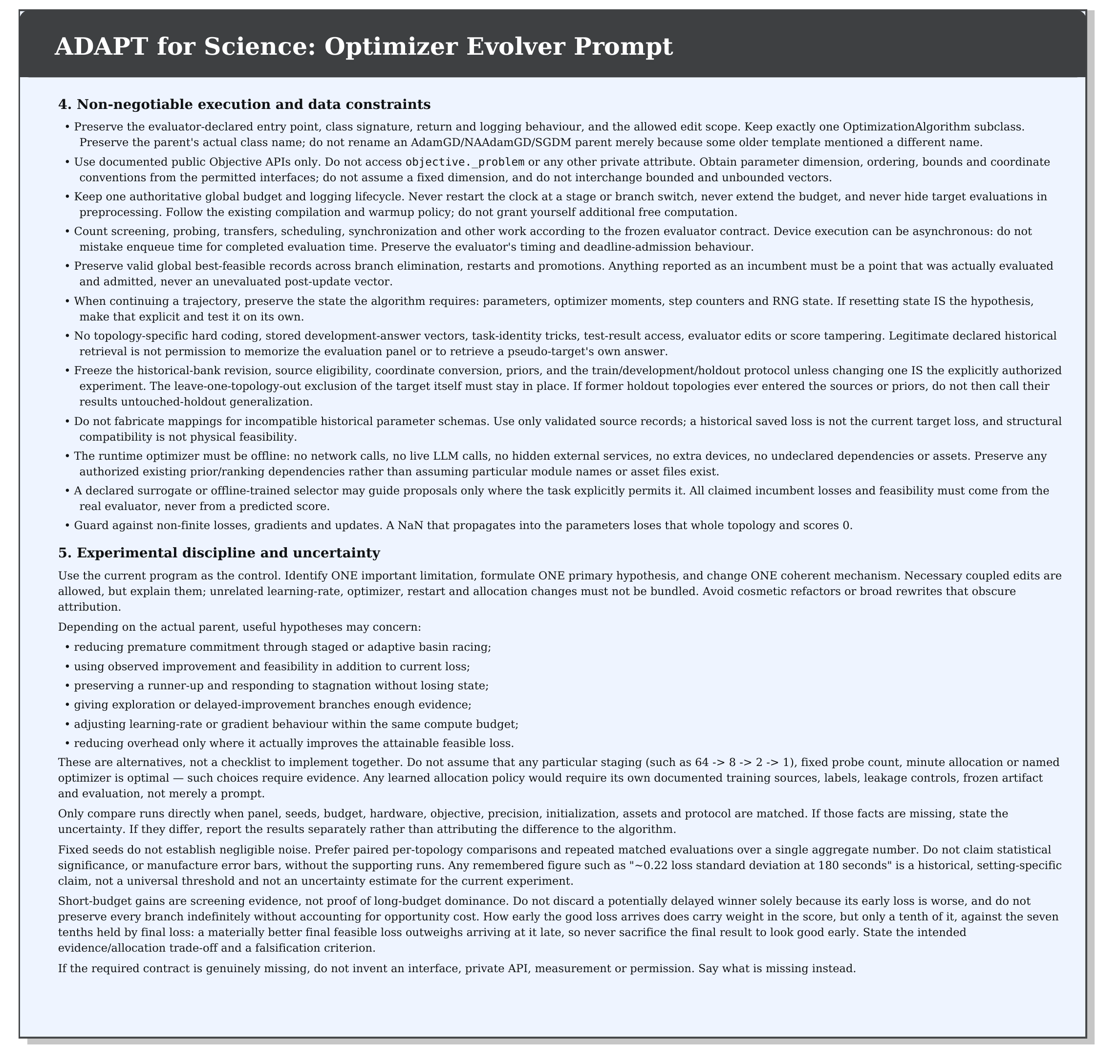}

    \caption*{
    \textbf{Figure~\ref{fig:supp_optimizer_prompt}, continued (page 2/3).}
    }
\end{figure}

\begin{figure}[p]
    \centering
    \includegraphics[
        page=3,
        width=\textwidth
    ]{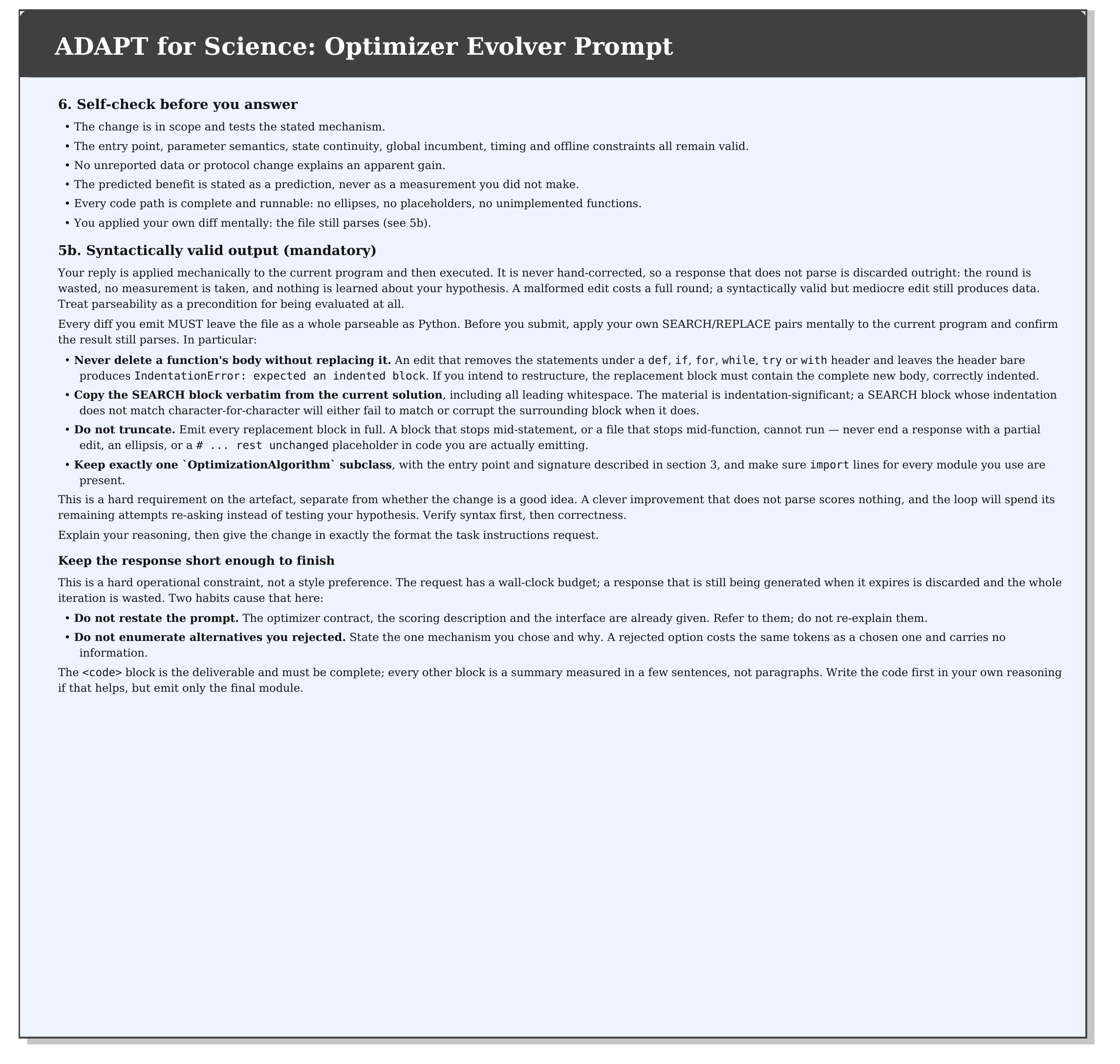}

    \caption*{
    \textbf{Figure~\ref{fig:supp_optimizer_prompt}, continued (page 3/3).}
    }
\end{figure}

\subsection{Anonymous Review and Artifact Release}
\label{app:anonymous_release}

The full prompts are included to make the agentic search procedure auditable
during anonymous review. The implementation source code and executable
artifacts are not distributed during the double-blind review period.
Upon acceptance, we plan to release the implementation, finalized
configuration files, and reproducibility artifacts.


\section{Experimental Protocol and Compute Accounting}
\label{app:experimental_setup}

This appendix summarizes the evaluation protocol and computational budget used
throughout the study. No additional experiments are introduced here; all
quantitative results correspond to the experiments reported in the main paper.

ADAPT separates \emph{method development} from \emph{held-out scientific
validation}. Candidate selection and optimization procedures are evolved using
development targets, after which the resulting procedures are frozen.
The eight held-out detector topologies are used only for final evaluation.

\subsection{Development and Held-Out Evaluation Regimes}
\label{app:development_validation_regimes}

The two stages operate under distinct computational regimes.

During development of the Selection Evolver, candidate selection procedures are
evaluated on the development panel with a maximum wall-clock budget of
\begin{equation}
    B_{\mathrm{dev}}
    =
    2~\mathrm{h}
\end{equation}
per target on an NVIDIA RTX~4090.

The final frozen ADAPT pipeline is evaluated on eight held-out detector
topologies. Each held-out target receives
\begin{equation}
    B_{\mathrm{test}}
    =
    4~\mathrm{h}
\end{equation}
on one NVIDIA H100 GPU.

The two regimes are intentionally kept separate. In particular, H100
validation outcomes are not returned to either the Selection Evolver or the
Optimizer Evolver and therefore cannot influence algorithm development.

The distinction is summarized in
Table~\ref{tab:supp_compute_protocol}.

\begin{table}[t]
    \centering
    \caption{
    \textbf{Development and held-out evaluation regimes.}
    Development feedback is obtained only from the development panel.
    The H100 evaluation is performed only after the final algorithms have
    been frozen.
    }
    \label{tab:supp_compute_protocol}
    \small
    \setlength{\tabcolsep}{6pt}
    \renewcommand{\arraystretch}{1.12}
    \begin{tabular}{llll}
        \toprule
        \textbf{Stage}
        & \textbf{Hardware}
        & \textbf{Budget}
        & \textbf{Used for evolution?} \\
        \midrule
        Selection development
        & RTX 4090
        & up to 2 h / target
        & Yes \\
        Optimizer development
        & Development hardware
        & fixed evaluator budget
        & Yes \\
        Held-out validation
        & NVIDIA H100
        & 4 h / target
        & No \\
        \bottomrule
    \end{tabular}
\end{table}

\subsection{Held-Out Validation Compute}
\label{app:h100_compute}

The held-out test set contains eight detector topologies. Since each target
receives a four-hour budget on one H100 GPU, the final ADAPT validation
requires
\begin{equation}
    C_{\mathrm{H100}}
    =
    8
    \times
    4
    =
    32
    \quad
    \text{H100 GPU-hours}.
    \label{eq:supp_h100_hours}
\end{equation}

Thus, the final held-out validation reported for ADAPT corresponds to
\begin{equation}
    \boxed{
        32~\text{H100 GPU-hours}
    }.
\end{equation}

When the eight targets are executed concurrently on eight GPUs, this
corresponds to approximately four hours of validator wall-clock time,
excluding scheduling and launch overhead. GPU-hours, rather than parallel
wall-clock time, are used for compute accounting.

\subsection{Development Compute Accounting}
\label{app:development_compute}

For the Selection Evolver, one complete evaluation of a candidate selection
procedure over the eight-target development panel, when all targets consume
their full two-hour budget, corresponds to at most
\begin{equation}
    C_{\mathrm{sel,eval}}
    =
    8\times2
    =
    16
    \quad
    \text{RTX 4090 GPU-hours}.
\end{equation}

The top-level selection-development pool contains
\begin{equation}
    4\times2\times2=16
\end{equation}
workflow--model--reference configurations, as described in
Appendix~\ref{app:selection_agent_pool}.

A full single sweep over these configurations would therefore have a nominal
upper-bound allocation of
\begin{equation}
    C_{\mathrm{sel,sweep}}
    =
    16
    \times
    16
    =
    256
    \quad
    \text{RTX 4090 GPU-hours},
    \label{eq:supp_selection_sweep_compute}
\end{equation}
if every configuration used the complete target-level budget.

Equation~\eqref{eq:supp_selection_sweep_compute} is a budget-accounting
quantity rather than a measured runtime: individual executions can terminate
before their maximum budget, and optimizer-evolution rounds constitute
additional development computation.

We therefore report H100 and RTX~4090 compute separately rather than converting
them into a single ``equivalent GPU-hour'' quantity. Such a conversion would
require assumptions about device-dependent simulator throughput.

\subsection{Single-Clock Evaluation}
\label{app:single_clock_evaluation}

All comparisons use matched random seeds, hardware, numerical precision, and
the same total wall-clock budget. For ADAPT, this budget covers both warmup
(candidate screening and staged probing) and subsequent optimization; no
additional time is allocated for warmup.

For selection, the charged time includes retrieval-side screening, permitted
probes, continuation of candidate trajectories, scheduler overhead, and the
remaining optimization trajectory. Time spent deciding where to search is
therefore unavailable to the final selected branch.

The final optimizer uses only the time remaining under the original shared
deadline; starting its logging does not reset the scientific budget. An
evaluation is eligible for reporting only if it is completed and admitted
before that deadline.

This convention prevents a method from obtaining additional scientific
evaluations through preprocessing, branch switching, asynchronous execution,
or post-deadline completion.

\subsection{Held-Out Isolation}
\label{app:protocol_isolation}

The eight held-out topologies are excluded from strategy development.
Their final optimization outcomes are not used for
\begin{itemize}
    \item selection-policy evolution;
    \item optimizer-program evolution;
    \item language-model selection;
    \item prompt revision;
    \item hyperparameter selection; or
    \item subsequent revision of ADAPT-Adam.
\end{itemize}

The target topology itself is necessarily available at test time because it
defines the scientific optimization problem. However, its held-out solution
and final validation outcome are not provided as development information.

The historical retrieval procedure also excludes the target itself when
constructing its 64-design candidate pool.

\subsection{What Is and Is Not Trained}
\label{app:what_is_trained}

ADAPT does not train a neural surrogate of the detector simulator for the
reported optimization results.

The computationally learned objects are instead the \emph{search procedures}:
the Selection Evolver develops a target-selection policy and the Optimizer
Evolver develops an executable numerical optimization algorithm.

Once these procedures are frozen, held-out detector optimization consists of
direct execution against the differentiable physics simulator. No language
model is fine-tuned during the held-out runs, and no online LLM inference is
required by the final numerical optimizer.


\section{Complete Held-Out Results}
\label{app:complete_results}

For completeness, this appendix consolidates the per-topology results reported
in the main paper. Lower loss is better in all tables.

The reported mean and standard deviation are computed across the eight
held-out detector topologies:
\begin{equation}
    \overline{L}
    =
    \frac{1}{8}
    \sum_{i=1}^{8}L_i,
\end{equation}
and
\begin{equation}
    \sigma_L
    =
    \sqrt{
        \frac{1}{8}
        \sum_{i=1}^{8}
        \left(
            L_i-\overline{L}
        \right)^2
    }.
    \label{eq:supp_cross_topology_std}
\end{equation}

Accordingly, the reported standard deviation describes variation
\emph{across detector topologies}; it is not a standard error and should not
be interpreted as repeated-seed uncertainty.

\subsection{Classical Optimizer Baselines}
\label{app:classical_full_results}

\begin{table}[t]
    \centering
    \caption{
    \textbf{Complete results for classical optimization baselines.}
    All entries are the best reported held-out loss under the matched
    scientific evaluation budget. Lower is better.
    }
    \label{tab:supp_classical_results}
    \scriptsize
    \setlength{\tabcolsep}{3.0pt}
    \renewcommand{\arraystretch}{1.12}
    \resizebox{\textwidth}{!}{
    \begin{tabular}{lccccccccc}
        \toprule
        \textbf{Method}
        & \textbf{T1}
        & \textbf{T2}
        & \textbf{T3}
        & \textbf{T4}
        & \textbf{T5}
        & \textbf{T6}
        & \textbf{T7}
        & \textbf{T8}
        & \textbf{Mean $\pm$ Std.} \\
        \midrule

        Adam
        & 0.423 & 0.887 & 1.833 & 0.332
        & 0.379 & 0.692 & 0.993 & 0.666
        & $0.776\pm0.458$ \\

        NAAdam
        & 0.423 & 0.696 & 0.423 & 0.334
        & 0.415 & 0.375 & 0.695 & 0.255
        & $0.452\pm0.150$ \\

        Lion
        & 4.358 & 3.841 & 2.028 & 3.028
        & 2.004 & 2.641 & 2.663 & 2.325
        & $2.861\pm0.793$ \\

        SGDM
        & 4.151 & 0.445 & 0.468 & 0.473
        & 0.581 & 0.448 & 0.196 & 0.415
        & $0.897\pm1.234$ \\

        Sophia
        & 0.506 & 0.412 & 0.627 & 1.749
        & 0.653 & 0.574 & 0.430 & 1.659
        & $0.826\pm0.514$ \\

        Adan
        & 1.404 & 1.321 & 1.376 & 2.783
        & 2.212 & 0.673 & 1.355 & 1.387
        & $1.564\pm0.602$ \\

        \midrule
        \textbf{ADAPT}
        & \textbf{0.311}
        & \textbf{0.408}
        & \textbf{0.075}
        & \textbf{0.228}
        & \textbf{0.150}
        & \textbf{0.139}
        & \textbf{0.163}
        & \textbf{0.155}
        & $\mathbf{0.204\pm0.101}$ \\

        \bottomrule
    \end{tabular}
    }
\end{table}

\subsection{General-Purpose Agentic Evolution Baselines}
\label{app:agentic_full_results}

\begin{table}[t]
    \centering
    \caption{
    \textbf{Complete results for general-purpose agentic evolution baselines.}
    Lower is better.
    }
    \label{tab:supp_agentic_results}
    \small
    \setlength{\tabcolsep}{5pt}
    \renewcommand{\arraystretch}{1.12}
    \begin{tabular}{lccccc}
        \toprule
        \textbf{Target}
        & \textbf{GEPA}
        & \textbf{OpenEvolve}
        & \textbf{ShinkaEvolve}
        & \textbf{ASI}
        & \textbf{ADAPT} \\
        \midrule
        T1 & 0.557 & 0.707 & 0.561 & 0.725 & \textbf{0.311} \\
        T2 & 0.410 & 0.492 & 0.440 & 0.483 & \textbf{0.408} \\
        T3 & 0.480 & 0.449 & 0.674 & 0.624 & \textbf{0.075} \\
        T4 & 1.218 & 0.261 & 2.632 & 0.653 & \textbf{0.228} \\
        T5 & 0.379 & 0.665 & 0.366 & 1.309 & \textbf{0.150} \\
        T6 & 0.435 & 0.426 & 0.210 & 0.450 & \textbf{0.139} \\
        T7 & 0.475 & 0.439 & 0.643 & 1.681 & \textbf{0.163} \\
        T8 & 0.365 & 0.954 & 0.402 & 1.067 & \textbf{0.155} \\
        \midrule
        Mean $\pm$ Std.
        & $0.540\pm0.263$
        & $0.549\pm0.202$
        & $0.741\pm0.729$
        & $0.874\pm0.410$
        & $\mathbf{0.204\pm0.101}$ \\
        \bottomrule
    \end{tabular}
\end{table}

\subsection{Selection and Optimizer Ablations}
\label{app:ablation_full_results}

The existing ablations isolate three components of the full system.

\begin{itemize}
    \item \textbf{Direct+GEPA} bypasses adaptive start selection and applies
    the agentic optimizer pipeline from a single retrieved initialization.

    \item \textbf{Fixed+Adam} retains the selected initialization but replaces
    the evolved optimizer with fixed Adam.

    \item \textbf{No-Prompt} removes the science-specific guidance while
    retaining the corresponding agentic evolution pipeline.
\end{itemize}

\begin{table}[t]
    \centering
    \caption{
    \textbf{Complete ablation results.}
    Lower is better. Mean $\pm$ standard deviation is computed across the
    eight held-out topologies.
    }
    \label{tab:supp_ablation_results}
    \small
    \setlength{\tabcolsep}{5pt}
    \renewcommand{\arraystretch}{1.12}
    \begin{tabular}{lcccc}
        \toprule
        \textbf{Target}
        & \textbf{Direct+GEPA}
        & \textbf{Fixed+Adam}
        & \textbf{No-Prompt}
        & \textbf{ADAPT} \\
        \midrule

        T1 & 0.416 & 0.320 & 0.424 & \textbf{0.311} \\
        T2 & 0.986 & \textbf{0.371} & 0.443 & 0.408 \\
        T3 & 0.373 & 0.180 & 0.145 & \textbf{0.075} \\
        T4 & 0.476 & 0.246 & 0.287 & \textbf{0.228} \\
        T5 & 0.164 & 0.200 & 0.264 & \textbf{0.150} \\
        T6 & \textbf{0.138} & 0.143 & 0.148 & 0.139 \\
        T7 & 0.413 & 0.415 & 0.499 & \textbf{0.163} \\
        T8 & 0.421 & 0.172 & 0.210 & \textbf{0.155} \\

        \midrule

        Mean $\pm$ Std.
        & $0.423\pm0.243$
        & $0.256\pm0.094$
        & $0.303\pm0.128$
        & $\mathbf{0.204\pm0.101}$ \\

        \bottomrule
    \end{tabular}
\end{table}

These ablations show that no single component explains the complete result.
Adaptive initialization improves robustness to the starting region,
optimizer evolution improves the subsequent local search relative to fixed
Adam, and science-specific guidance improves the agentic evolution procedure.

The per-target results also show that these improvements are not strictly
uniform. Fixed+Adam obtains a lower loss than full ADAPT on T2, while
Direct+GEPA is marginally lower on T6. We retain these cases rather than
reporting only aggregate improvements.


\section{Complete Held-Out Detector Topology Atlas}
\label{app:topology_atlas}

This appendix provides the complete visual comparison for all eight held-out
detector topologies.

For each target, we show
\begin{enumerate}
    \item the initial detector topology;
    \item the solution obtained by the standard Adam baseline; and
    \item the final solution obtained by ADAPT.
\end{enumerate}

The optical topology is fixed during continuous optimization. Visual
differences in the optimized panels arise from changes in the continuous
physical parameters. Components whose optimized contribution becomes
negligible are displayed as vacuum inputs where applicable.

\subsection{Held-Out Topology Definitions}
\label{app:heldout_topology_definitions}

Table~\ref{tab:supp_topology_ids} maps the shorthand target identifiers used
throughout the paper to their topology strings.

\begin{table}[t]
    \centering
    \caption{
    \textbf{Eight held-out detector topologies.}
    The first string specifies the $3\times3$ internal optical layout and the
    second string specifies the boundary components.
    }
    \label{tab:supp_topology_ids}
    \small
    \setlength{\tabcolsep}{9pt}
    \renewcommand{\arraystretch}{1.12}
    \begin{tabular}{lll}
        \toprule
        \textbf{Target}
        & \textbf{Internal topology}
        & \textbf{Boundary topology} \\
        \midrule
        T1 & \texttt{DDDBBBDDC} & \texttt{LSLLSSSLSHSS} \\
        T2 & \texttt{EAFBHDGGG} & \texttt{LLSSSSDLSLSS} \\
        T3 & \texttt{FADGHCFEC} & \texttt{LLLLLLLLLLLH} \\
        T4 & \texttt{FGHHBGEFE} & \texttt{DLLLLLLLLLLS} \\
        T5 & \texttt{HAFHBEBFB} & \texttt{SLLLDLSLSLLL} \\
        T6 & \texttt{HHCDHEHBF} & \texttt{SDSSSSSLSSSS} \\
        T7 & \texttt{HHECHECAB} & \texttt{LLSSSSHLSLSS} \\
        T8 & \texttt{HHFFGFGBH} & \texttt{SSLLSSLSSLDL} \\
        \bottomrule
    \end{tabular}
\end{table}

\subsection{Eight-Target Visual Comparison}
\label{app:all_topology_visuals}

\begin{figure}[!p]
    \centering

    \vspace*{-0.8cm}

    \includegraphics[
        width=0.9\textwidth,
        height=0.9\textheight,
        keepaspectratio
    ]{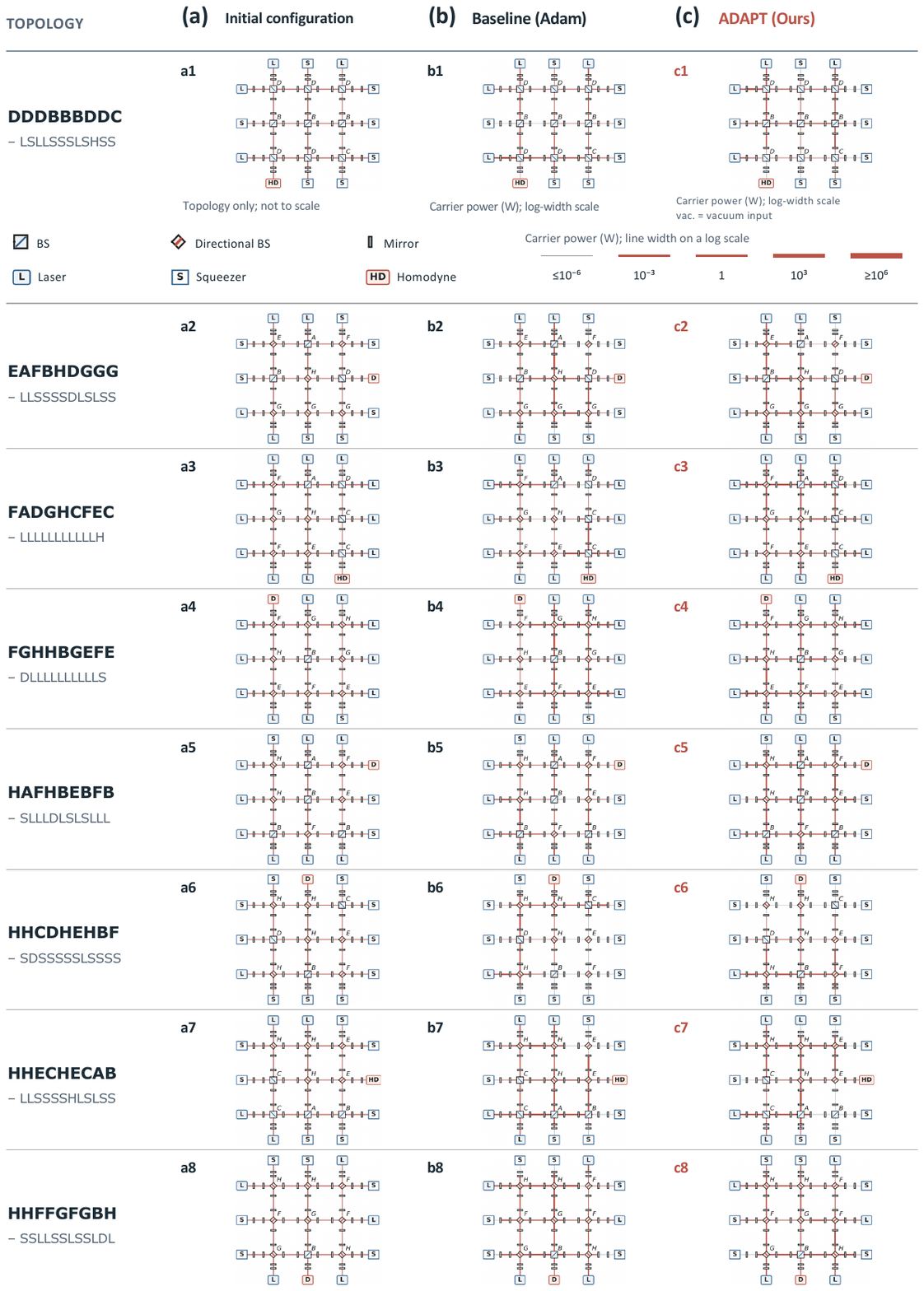}

    \vspace{-10pt}

    \caption{
    \textbf{Complete comparison across the eight held-out detector
    topologies.}
    Each row corresponds to one target.
    Column (a) shows the initial detector configuration,
    column (b) shows the optimized standard-Adam baseline, and
    column (c) shows the final ADAPT design.
    Carrier power is represented by line width on a logarithmic scale.
    Components with effectively vanishing optimized contribution are shown
    as vacuum inputs where applicable.
    }
    \label{fig:supp_all_topologies}
\end{figure}

Figure~\ref{fig:supp_all_topologies} contains the complete $8\times3$
comparison.

The atlas emphasizes that the performance difference is not restricted to a
single detector configuration. Across the held-out panel, the optimizer acts
on different continuous parameter landscapes while preserving the specified
discrete topology and physical simulator.


\section{Reproducibility, Reporting Conventions, and Scope}
\label{app:reproducibility}

This appendix summarizes the conventions required to interpret and reproduce
the reported results. The submission is under double-blind review; source code
is therefore not linked to an identity-bearing public repository during the
review period. We plan to release the implementation and associated
reproducibility artifacts after acceptance.

\subsection{Reproduction Boundary}
\label{app:reproduction_boundary}

A complete reproduction of one held-out ADAPT run requires the following
components:

\begin{enumerate}
    \item the frozen detector topology and continuous parameter schema;
    \item the differentiable physics simulator and native scientific objective;
    \item the historical design library;
    \item the target-aware retrieval and semantic-transfer procedure;
    \item the frozen Selection Evolver policy;
    \item the frozen ADAPT-Adam optimizer;
    \item the global wall-clock budget and deadline-admission rules; and
    \item the held-out reporting procedure.
\end{enumerate}

The complete mathematical definitions of the scientific objective and
continuous parameterization are given in
Appendix~\ref{app:scientific_objective}.
Historical retrieval is described in
Appendix~\ref{app:historical_library},
selection in Appendix~\ref{app:selection_evolution},
and the final optimizer in Appendix~\ref{app:optimizer_evolution}.

\subsection{Information Available at Held-Out Runtime}
\label{app:runtime_information}

During held-out execution, the frozen numerical procedure may access only
information naturally supplied by the scientific optimization environment,
including

\begin{itemize}
    \item the target topology;
    \item the current parameter vector;
    \item the native objective and its gradient;
    \item physical-feasibility information exposed by the evaluator;
    \item the best evaluated incumbent;
    \item the number of completed evaluations; and
    \item the remaining wall-clock budget.
\end{itemize}

The runtime procedure does not access
\begin{itemize}
    \item held-out answer vectors;
    \item previously recorded held-out final losses;
    \item topology-specific manually encoded solutions;
    \item evaluator-private state;
    \item live language-model inference;
    \item external network services; or
    \item post-deadline scientific evaluations.
\end{itemize}

\subsection{Best-Feasible Reporting Rule}
\label{app:best_feasible_reporting}

All final results use the best physically feasible point that was actually
evaluated and admitted before the corresponding deadline.

For a target $\tau$ and admitted evaluation set
$\mathcal{E}_{\tau}(B)$,
the reported loss is
\begin{equation}
    L_{\tau}^{\mathrm{reported}}
    =
    \min_{
        x\in
        \mathcal{E}_{\tau}(B)
        \cap
        \mathcal{F}_{\tau}
    }
    L_{\tau}(x).
    \label{eq:supp_reported_loss}
\end{equation}

An unevaluated post-update vector, predicted surrogate value, or result
completed after the budget expires is not eligible for reporting.

\subsection{Interpretation of Variability}
\label{app:variability_reporting}

Unless explicitly stated otherwise, reported values of the form
\begin{equation}
    \text{mean}\pm\text{standard deviation}
\end{equation}
refer to variation across the eight held-out detector topologies.

They do not constitute
\begin{itemize}
    \item repeated-seed uncertainty;
    \item a confidence interval;
    \item a standard error of the mean; or
    \item evidence of statistical significance.
\end{itemize}

We therefore base the main conclusions on the complete per-topology results
in addition to aggregate means.

\subsection{Scope and Limitations}
\label{app:supp_scope}

The present evaluation establishes ADAPT in one scientific domain:
differentiable optimization of gravitational-wave detector designs with
reusable historical experience and explicit physical constraints.

Several limitations should therefore be kept in mind.

First, the effectiveness of target-aware retrieval depends on the coverage and
quality of the historical experience library. Its benefit may decrease for
targets that are far outside the observed design distribution.

Second, evolving both initialization and optimization procedures introduces
development-time computational overhead relative to applying a single fixed
optimizer.

Third, the present validator is simulation-based. Although the simulator
provides a controlled differentiable scientific environment, the current
results do not establish performance under real experimental noise,
hardware imperfections, or non-differentiable laboratory feedback.

Finally, ADAPT-Adam is an empirically discovered finite-budget optimizer.
The present work does not claim a global convergence guarantee for the evolved
update rule.

These limitations define the scope of the reported evidence rather than
additional unreported experiments.

\ifdefined\FROMMAIN
\else
    \end{document}
\fi

\end{document}